\documentclass[runningheads]{llncs}
\usepackage{graphicx}
\usepackage{amsmath,amssymb} 
\usepackage{color}
\usepackage{xcolor}
\usepackage{booktabs}
\usepackage{multirow}
\usepackage{caption}
\usepackage{footmisc}
\usepackage[compress]{cite}
\usepackage[breaklinks=true,colorlinks=false,bookmarks=false]{hyperref} 

\newcommand{\etal}{\textit{et al}.}

\DeclareMathOperator*{\argmin}{\arg\!\min}

\begin{document}
    
\pagestyle{headings}
\mainmatter

\title{BodyNet: Volumetric Inference of \\ 3D Human Body Shapes}

\titlerunning{BodyNet: Volumetric Inference of 3D Human Body Shapes}

\authorrunning{Varol, Ceylan, Russell, Yang, Yumer, Laptev, Schmid}

\author{G\"{u}l Varol\textsuperscript{$1$,*} \qquad  Duygu Ceylan\textsuperscript{$2$}
    \qquad  Bryan Russell\textsuperscript{$2$} \qquad Jimei Yang\textsuperscript{$2$} \\ 
    Ersin Yumer\textsuperscript{$2$,\ddag} \qquad  Ivan Laptev\textsuperscript{$1$,*} \qquad
      Cordelia Schmid\textsuperscript{$1$,\dag}}
\institute{\textsuperscript{$1$}Inria, France \qquad \qquad \textsuperscript{$2$}Adobe Research, USA}

\maketitle

\footnotetext[1]{\'{E}cole normale sup\'{e}rieure, Inria, CNRS, 
    PSL Research University, Paris, France}
\footnotetext[2]{Univ. Grenoble Alpes, Inria, CNRS, INPG, LJK, Grenoble, France}
\footnotetext[3]{Currently at Argo AI, USA. This work was performed while EY was at Adobe.}

\begin{abstract}

Human shape estimation is an important task for video editing, animation and fashion industry. Predicting 3D human body shape from natural images, however, is highly challenging due to factors such as variation in human bodies, clothing and viewpoint. Prior methods addressing this problem typically attempt to fit parametric body models with certain priors on pose and shape. In this work we argue for an alternative representation and propose BodyNet, a neural network for direct inference of volumetric body shape from a single image. BodyNet is an end-to-end trainable network that benefits from (i) a volumetric 3D loss, (ii) a multi-view re-projection loss, and (iii) intermediate supervision of 2D pose, 2D body part segmentation, and 3D pose. Each of them results in performance improvement as demonstrated by our experiments. To evaluate the method, we fit the SMPL model to our network output and show state-of-the-art results on the SURREAL and Unite the People datasets, outperforming recent approaches. Besides achieving state-of-the-art performance, our method also enables volumetric body-part segmentation.
\end{abstract}

\section{Introduction}
\label{sec:introduction}
Parsing people in visual data is central to many applications including mixed-reality interfaces, animation, video editing and human action recognition.
Towards this goal, human 2D pose estimation has been significantly advanced by recent efforts~\cite{newell2016hourglass,wei2016cpm,pishchulin16cvpr,cao2017realtime}.
Such methods aim to recover 2D locations of body joints and provide a simplified geometric representation of the human body.
There has also been significant progress in 3D human pose estimation~\cite{Martinez2017simple,pavlakos2017volumetric,rogez2017lcrnet,zhou2017towards3d}. 
Many applications, however, such as virtual clothes try-on, video editing and re-enactment require accurate estimation of both 3D human {\em pose} and {\em shape}.

\begin{figure}[t]
	\begin{center}
	  \includegraphics[width=0.8\linewidth]{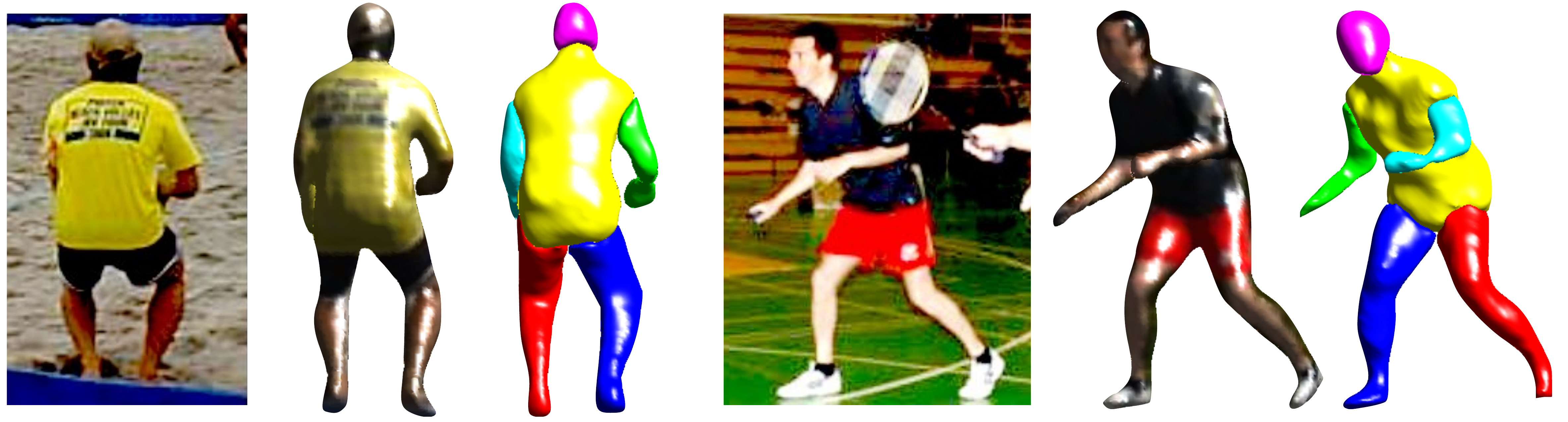}
	\end{center}
	\caption{Our BodyNet predicts a volumetric 3D human body shape and 3D body parts from a single image.
		We show the input image, the predicted human voxels, and the predicted part voxels.}
	\label{fig:teaser}
\end{figure}

3D human shape estimation has been mostly studied in controlled settings using specific sensors including multi-view capture~\cite{Leroy2017}, motion capture markers~\cite{loper2014mosh}, inertial sensors~\cite{marcard2017imu}, and 3D scanners~\cite{yang2016scan}.
In uncontrolled single-view settings 3D human shape estimation, however, has received little attention so far.
The challenges include the lack of large-scale training data, the high dimensionality of the output space, and the choice of suitable representations for 3D human shape.
Bogo \etal~\cite{Bogo2016smplify} present the first automatic method to fit a deformable body model to
an image but rely on accurate 2D pose estimation
and introduce hand-designed constraints enforcing elbows and knees to bend naturally.
Other recent methods~\cite{tan2017bmvc,tung2017selfsupervised,hmrKanazawa17} employ deformable human
body models such as SMPL~\cite{smpl2015} and regress model parameters with CNNs~\cite{krizhevsky2012,LeCun1989cnn}.
In this work, we compare to such approaches and show advantages.

The optimal choice of 3D representation for neural networks remains an open problem.
Recent work explores voxel~\cite{Maturana2015iros,Yan2016,yumer2016learning,Girdhar16b}, 
octree~\cite{ogn2017,Riegler2017OctNet,Wang2017ocnn,Riegler20173DV}, point cloud~\cite{su2017pointset,su2017pointnet,ppfnet18},
and surface~\cite{groueix2018} representations for modeling generic 3D objects.
In the case of human bodies, the common approach has been to regress parameters of
pre-defined human shape models~\cite{tan2017bmvc,tung2017selfsupervised,hmrKanazawa17}. 
However, the mapping between the 3D shape and parameters of deformable body models is highly nonlinear
and is currently difficult to learn.
Moreover, regression to a single set of parameters cannot represent multiple hypotheses and can be problematic in ambigous situations.
Notably, skeleton regression methods for 2D human pose estimation, e.g.,~\cite{Toshev:2014}, have recently been
overtaken by heatmap based methods~\cite{newell2016hourglass,wei2016cpm} enabling representation of multiple hypotheses.

\begin{figure}[t]
	\begin{center}
	  \includegraphics[width=\textwidth]{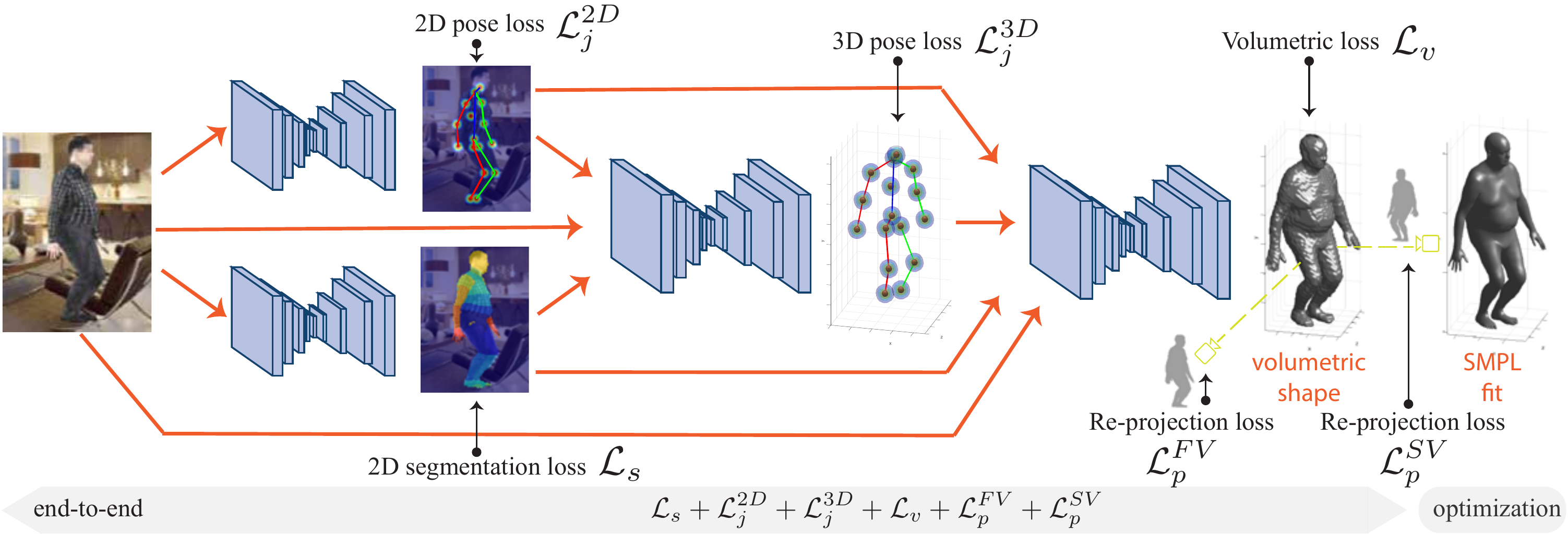}
	\end{center}
	\caption{BodyNet: End-to-end trainable network for
	  3D human body shape estimation. The input RGB image is first passed through subnetworks
	  for 2D pose estimation and 2D body part segmentation. These predictions, combined with the
	  RGB features, are fed to another network predicting 3D pose. All subnetworks are
	  combined to a final network to infer volumetric shape. The 2D pose, 2D segmentation and 3D pose
          networks are first pre-trained and then fine-tuned jointly for the task of
          volumetric shape estimation using multi-view re-projection losses.
          We fit the SMPL model to volumetric predictions for the purpose of evaluation.}
	\label{fig:pipeline}
\end{figure}

In this work we propose and investigate a volumetric representation for body shape estimation as illustrated in Fig.~\ref{fig:teaser}.
Our network, called BodyNet, generates likelihoods on the 3D occupancy grid of a person.
To efficiently train our network, we propose to regularize BodyNet with a set of auxiliary losses.
Besides the main volumetric 3D loss, BodyNet includes a multi-view re-projection loss and multi-task losses.
The multi-view re-projection loss, being efficiently approximated on voxel space (see Sec.~\ref{subsec:re-projection}), increases the importance of the boundary voxels.
The multi-task losses are based on the additional intermediate network supervision in terms of 2D pose, 2D body part
segmentation, and 3D pose.
The overall architecture of BodyNet is illustrated in Fig.~\ref{fig:pipeline}.

To evaluate our method, we fit the SMPL model~\cite{Bogo2016smplify}
to the BodyNet output and measure single-view 3D human shape
estimation performance in the recent SURREAL~\cite{varol2017surreal} and Unite the People~\cite{lassner2017up} datasets. 
The proposed BodyNet approach demonstrates state-of-the-art
performance and improves accuracy of recent methods. We show
significant improvements provided by the end-to-end training and
auxiliary losses of BodyNet. 
Furthermore, our method enables volumetric body-part segmentation.
BodyNet is fully-differentiable and could be used as a subnetwork in
future application-oriented methods targeting e.g., virtual cloth
change or re-enactment. 

In summary, this work makes several contributions.
First, we address single-view 3D human shape estimation and propose a volumetric representation for this task.
Second, we investigate several network architectures and propose an end-to-end trainable network BodyNet combining a multi-view re-projection loss together with intermediate network supervision in terms of 2D pose, 2D body part segmentation, and 3D pose.
Third, we outperform previous regression-based methods and demonstrate state-of-the art performance on two datasets for human shape estimation.
In addition, our network is fully differentiable and can provide volumetric body-part segmentation.

\section{Related work}
\label{sec:relatedwork}

\noindent{\bf  3D human body shape.}
While the problem of localizing 3D body joints has been well-explored in the past~\cite{h36m_pami,kostrikov2014,Martinez2017simple,pavlakos2017volumetric,rogez2017lcrnet,Yasin2016,zhou2017towards3d,rogez2016nips},
3D human {\em shape} estimation from a single image has received limited attention and remains a challenging problem. Earlier work~\cite{Balan:CVPR:2007,guan2009} proposed to optimize pose and shape parameters of the 3D deformable body model SCAPE~\cite{scape}. 
More recent methods use the SMPL~\cite{smpl2015} body model that again represents the 3D
shape as a function of pose and shape parameters. Given
such a model and an input image, Bogo \etal~\cite{Bogo2016smplify} present the
optimization method SMPLify estimating model parameters from a fit to 2D joint locations.
Lassner \etal~\cite{lassner2017up} extend this approach by incorporating
silhouette information as additional guidance and improves the optimization performance
by densely sampled 2D points.
Huang~\etal~\cite{MuVS:3DV:2017} extend SMPLify for multi-view video sequences
with temporal priors. Similar temporal constraints have been used in~\cite{gcpr2017}.
Rhodin~\etal~\cite{rhodin2016} use a sum-of-Gaussians volumetric
representation together with contour-based refinement and successfully demonstrate
human shape recovery from multi-view videos with optimization techniques.
Even though such methods show compelling results, inherently they are limited by the
quality of the 2D detections they use and depend on priors both on pose and shape
parameters to regularize the highly complex and costly optimization process.

Deep neural networks provide an alternative approach that can be expected to learn
appropriate priors automatically from the data. Dibra \etal~\cite{Dibra2016a} present one of the
first approaches in this direction and train a CNN to estimate the 3D shape parameters from silhouettes, but assume a frontal input view.
More recent approaches~\cite{tan2017bmvc,tung2017selfsupervised,hmrKanazawa17} train neural networks to 
predict the SMPL body parameters from an input image.
Tan~\etal~\cite{tan2017bmvc} design an encoder-decoder architecture that is trained on silhouette prediction and indirectly
regresses model parameters at the bottleneck layer.
Tung~\etal~\cite{tung2017selfsupervised} operate on two consecutive video frames and learn parameters by integrating re-projection
loss on the optical flow, silhouettes and 2D joints.
Similarly, Kanazawa~\etal~\cite{hmrKanazawa17} predict parameters with re-projection loss on the 2D joints
and introduce an adversary whose goal is to distinguish unrealistic human body shapes.

Even though parameters of deformable body models provide a low-dimensional
embedding of the 3D shape, predicting such parameters with a network requires learning a highly non-linear mapping.
In our work we opt for an alternative volumetric representation that has 
shown to be effective for generic 3D objects~\cite{Yan2016} and 
faces~\cite{jackson2017vrn}. 
The approach of~\cite{Yan2016} operates on low-resolution grayscale images for a few rigid object
categories such as chairs and tables.
We argue that human bodies are more challenging due to significant non-rigid deformations.
To accommodate for such deformation, we use segmentation and 3D pose as proxy to 3D shape in addition 
to 2D pose~\cite{jackson2017vrn}. Conditioning our 3D shape estimation on a given 3D pose,
the network focuses on the more complicated problem of shape deformation.
Furthermore, we regularize our voxel predictions with additional re-projection loss, perform
end-to-end multi-task training with intermediate supervision and obtain volumetric body part segmentation. 

Others have studied predicting 2.5D projections of human bodies.
DenseReg~\cite{guler2016densereg} and DensePose~\cite{Guler2018DensePose}
estimate image-to-surface correspondences,
while \cite{varol2017surreal} outputs quantized
depth maps for SMPL bodies. Differently from these methods, our approach generates a full 3D body reconstruction.

\noindent{\bf Multi-task neural networks.}
Multi-task networks are well-studied. A common approach is to output
multiple related tasks at the very end of the neural network architecture.
Another, more recently explored alternative is to stack multiple
subnetworks and provide guidance with {\em intermediate supervision}.
Here, we only cover related works that
employ the latter approach.
Guiding CNNs with relevant cues has shown improvements for a number of tasks.
For example, 2D facial landmarks have shown useful guidance for 3D
face reconstruction~\cite{jackson2017vrn} and similarly optical flow
for action recognition~\cite{simonyan2014}. However, these methods
do not perform joint training. Recent work of \cite{luvizonCVPR2018}
jointly learns 2D/3D pose together with action recognition. Similarly,
\cite{popa2017deepmultitask} trains for 3D pose with intermediate
tasks of 2D pose and segmentation.
With this motivation, we make use of 2D pose, 2D human body part 
segmentation, and 3D pose, that provide cues for 3D human
{\em shape} estimation. Unlike~\cite{popa2017deepmultitask},
3D pose becomes an auxiliary task for our final 3D shape task.
In our experiments, we
show that training with a joint loss on all these tasks increases the performance of
all our subnetworks (see Appendix~\ref{app:subsec:multitaskeffect}).

\section{BodyNet}
\label{sec:approach}
BodyNet predicts 3D human body shape from a single image and is composed of four subnetworks
trained first independently, then jointly to predict 2D pose, 2D  body part segmentation,
3D pose, and 3D shape (see Fig.~\ref{fig:pipeline}). Here, we first discuss the details of
the volumetric representation for body shape (Sec.~\ref{subsec:volumetric}). Then, we describe
the multi-view re-projection loss (Sec.~\ref{subsec:re-projection}) and the multi-task training with the
intermediate representations (Sec.~\ref{subsec:multi-task}). Finally, we formulate our model fitting
procedure (Sec.~\ref{subsec:fitting}).

\subsection{Volumetric inference for 3D human shape}
\label{subsec:volumetric}
For 3D human body shape, we propose to use a voxel-based representation.
Our shape estimation subnetwork outputs the 3D shape
represented as an occupancy map defined on a fixed resolution voxel grid.
Specifically, given a 3D body, we define a 3D voxel grid roughly centered at the root joint,
(i.e., the hip joint) 
where each voxel inside the body is marked as occupied.
We voxelize the ground truth meshes (i.e., SMPL) into a fixed 
resolution grid using binvox~\cite{nooruddin03,binvox}.
We assume orthographic projection and rescale the volume such that the $xy$-plane is aligned with the 2D segmentation mask
to ensure spatial correspondence with the input image.
After scaling, the body is centered on the $z$-axis and the remaining areas are padded with zeros.

Our network minimizes the binary cross-entropy loss 
after applying the sigmoid function on the network output similar to~\cite{jackson2017vrn}:
\begin{equation}
\mathcal{L}_v = \sum_{x=1}^{W} \sum_{y=1}^{H} \sum_{z=1}^{D} V_{xyz}\log \hat{V}_{xyz}+(1-V_{xyz})\log (1-\hat{V}_{xyz}),
\end{equation}
where $V_{xyz}$ and $\hat{V}_{xyz}$ denote the ground truth value and the predicted sigmoid output for a voxel,
respectively.
Width ($W$), height ($H$) and depth ($D$) are 128 in our experiments.
We observe that this resolution captures sufficient details.

The loss $\mathcal{L}_v$ is used to perform foreground-background segmentation of the voxel grid.
We further extend this formulation to perform 3D body part segmentation with a multi-class cross-entropy
loss. We define 6 parts (head, torso, left/right leg, left/right arm) and learn 7-class classification
including the background. The weights for this network are initialized by the shape network
by copying the output layer weights for each class. This simple extension allows the network to directly
infer 3D body parts without going through the costly SMPL model fitting.

\subsection{Multi-view re-projection loss on the silhouette}
\label{subsec:re-projection}
Due to the complex articulation of the human body, one major challenge in inferring the volumetric body shape 
is to ensure high confidence predictions across the whole body. We often observe 
that the confidences on the limbs away from the body center
tend to be lower (see Fig.~\ref{fig:projection}). 
To address this problem, we employ additional 2D re-projection losses 
that increase the importance of the boundary voxels. 
Similar losses have been employed for rigid objects by~\cite{zhu2017reprojection,drcTulsiani17} in the absence
of 3D labels and by~\cite{Yan2016} as additional regularization. In our case, we show that the multi-view re-projection term is critical,
particularly to obtain good quality reconstruction of body limbs.
Assuming orthographic projection, 
the front view projection, $\hat{S}^{FV}$,
is obtained by projecting the volumetric grid to the image with the {\em max} operator along the $z$-axis~\cite{zhu2017reprojection}. Similarly, we define 
$\hat{S}^{SV}$ as the {\em max} along the $x$-axis:
\begin{equation}
\hat{S}^{FV}(x,y) = \max_z \hat{V}_{xyz} \quad\text{and}\quad \hat{S}^{SV}(y,z) = \max_x \hat{V}_{xyz}.
\end{equation}
The true silhouette, $S^{FV}$, is defined by the ground truth 2D body part segmentation provided by 
the datasets. We obtain the ground truth side view silhouette from the voxel 
representation that we computed from the ground truth 3D mesh: ${S}^{SV}(y,z) = \max_x {V}_{xyz}$. We note 
that our voxels remain slightly larger than the original mesh due to the voxelization step that marks 
every voxel that intersects with a face as occupied. We define a binary cross-entropy loss per view 
as follows: 
\small
\begin{align}
& \mathcal{L}^{FV}_p = \sum_{x=1}^{W} \sum_{y=1}^{H} S(x,y)\log \hat{S}^{FV}(x,y)+(1-S(x,y))\log(1-\hat{S}^{FV}(x,y)), \\
& \mathcal{L}^{SV}_p = \sum_{y=1}^{H} \sum_{z=1}^{D} S(y,z)\log \hat{S}^{SV}(y,z)+(1-S(y,z))\log(1-\hat{S}^{SV}(y,z)).
\end{align}
\normalsize
We train the shape estimation network initially with $\mathcal{L}_v$. Then, we continue training
with a combined loss:
$\lambda_v\mathcal{L}_v  + \lambda^{FV}_p\mathcal{L}^{FV}_p + \lambda^{SV}_p\mathcal{L}^{SV}_p$, Sec.~\ref{subsec:multi-task} gives details on how to set the relative weighting of the losses. 
Sec.~\ref{subsec:3Dshape} demonstrates experimentally the benefits of the multi-view re-projection loss. 

\subsection{Multi-task learning with intermediate supervision}
\label{subsec:multi-task}
The input to the 3D shape estimation subnetwork is
composed by combining RGB, 2D pose, segmentation, and 3D pose predictions.
Here, we present the subnetworks used to predict these intermediate representations
and detail our multi-task learning procedure.
The architecture
for each subnetwork
is based on a stacked hourglass network~\cite{newell2016hourglass}, where the
output is over a spatial grid and is, thus, convenient for pixel-
and voxel-level tasks as in our case. 

\noindent{\bf 2D pose.}
Following the work of Newell \etal~\cite{newell2016hourglass}, we use a heatmap representation of 2D pose.
We predict one heatmap for each body joint where a Gaussian with fixed variance is centered at the
corresponding image location of the joint. The final joint locations are identified as 
the pixel indices with the maximum value over each output channel.
We use the first two stacks of an hourglass network to map RGB features
$3\times256\times256$ to 2D joint heatmaps
$16\times64\times64$
as in~\cite{newell2016hourglass} and predict $16$ body joints.
The mean-squared error between the ground truth
and predicted 2D heatmaps is $\mathcal{L}^{2D}_{j}$. 

\noindent{\bf 2D part segmentation.}
Our body part segmentation network is adopted from~\cite{varol2017surreal} 
and is trained on the SMPL~\cite{smpl2015} anatomic parts defined by~\cite{varol2017surreal}. The architecture is similar to the
2D pose network and again the first two stacks are used. The network predicts one heatmap per body part
given the input RGB image, which results in an output resolution of
$15\times64\times64$ for 15 body parts. The spatial cross-entropy loss is denoted
with $\mathcal{L}_{s}$.

\noindent{\bf 3D pose.}
Estimating the 3D joint locations from a single image is an inherently ambiguous
problem. To alleviate some uncertainty, we assume that the camera intrinsics are known and predict the 3D pose
in the camera coordinate system. Extending the notion of
2D heatmaps to 3D, we represent 3D joint locations with 3D Gaussians defined on a voxel grid as in~\cite{pavlakos2017volumetric}. 
For each joint, the network predicts a fixed-resolution volume with a
single 3D Gaussian centered at the joint location. The $xy-$dimensions of this grid are aligned with the image
coordinates, and hence the 2D joint locations, while the $z$ dimension represents the depth. We assume this voxel grid is
aligned with the 3D body such that the root joint corresponds to the center of the 3D volume. 
We determine a reasonable depth range in which a human body can fit (roughly $85$cm in our experiments) and quantize this range into 19 bins.  
We define the overall resolution of the 3D grid to be 
$64\times64\times19$, i.e., four times smaller in spatial resolution compared
to the input image as is the case for the 2D pose and segmentation networks. 
We define one such grid per body joint and regress with mean-squared error $\mathcal{L}^{3D}_j$.

The 3D pose estimation network consists of another two stacks.
Unlike 2D pose and segmentation, the 3D pose network
takes multiple modalities as input, all spatially aligned with the output of the network. Specifically, we concatenate 
RGB channels with the heatmaps corresponding to 2D joints
and body parts. We upsample the heatmaps to  match the RGB 
resolution, thus the input resolution becomes $(3+16+15)\times256\times256$.
While 2D pose provides a significant
cue for the $x,y$ joint locations, some of the depth information is implicitly contained in body part segmentation since unlike a silhouette,
occlusion relations among individual body parts provide strong 3D cues. For example a
discontinuity on the torso segment caused by an occluding arm segment implies the arm is in front
of the torso. In Appendix~\ref{app:subsec:3Dpose}, we provide comparisons of 3D pose prediction with and without using this 
additional information.

\noindent{\bf Combined loss and training details.} The subnetworks are initially trained independently with individual losses, then fine-tuned jointly with a combined loss:
\begin{equation}
\mathcal{L} =   \lambda^{2D}_j\mathcal{L}^{2D}_j + \lambda_s\mathcal{L}_s + \lambda^{3D}_j\mathcal{L}^{3D}_j + \lambda_v\mathcal{L}_v + \lambda^{FV}_p\mathcal{L}^{FV}_p + \lambda^{SV}_p\mathcal{L}^{SV}_p.
\end{equation}
The weighting coefficients are set such that the average gradient of each loss across parameters is at the same scale
at the beginning of fine-tuning. With this rule, we set
$ ( \lambda^{2D}_j, \lambda_s,  \lambda^{3D}_j,  \lambda_v, \lambda^{FV}_p,  \lambda^{SV}_p ) \propto ( 10^7, 10^3, 10^6, 10^1, 1, 1 ) $
and make the sum of the weights equal to one.
We set these weights on the SURREAL dataset and use the same values in all experiments.
We found it important to apply this balancing so that the network does not forget
the intermediate tasks, but improves the performance of all tasks at the same time.

When training our full network, see Fig.~\ref{fig:pipeline}, we proceed as follows:
(i) we train 2D pose and segmentation; (ii) we train 3D pose with fixed 2D pose and segmentation network weights;
(iii) we train 3D shape network with all the preceding network weights fixed;
(iv) then, we continue training the shape network with additional re-projection losses;
(v) finally, we perform end-to-end fine-tuning on all network weights with the combined loss.

\noindent{\bf Implementation details.} 
Each of our subnetworks consists of two stacks to keep a reasonable computational cost.
We take the first two stacks of the 2D pose network trained on the
MPII dataset~\cite{andriluka14mpii} with 8 stacks~\cite{newell2016hourglass}.
Similarly, the segmentation network is trained on the SURREAL dataset
with 8 stacks~\cite{varol2017surreal} and the first two stacks are used.
Since stacked hourglass networks involve intermediate supervision~\cite{newell2016hourglass},
we can use only part of the network by sacrificing slight performance.
The weights for 3D pose and 3D shape networks are randomly initialized and
trained on SURREAL with two stacks.
Architectural details are given in Appendix~\ref{app:subsec:archdetails}.
SURREAL~\cite{varol2017surreal}, being a large-scale dataset, provides pre-training for the UP dataset~\cite{lassner2017up}
where the networks converge relatively faster. Therefore, we fine-tune
the segmentation, 3D pose, and 3D shape networks on UP from those
pre-trained on SURREAL.
We use RMSprop~\cite{Tieleman2012} algorithm with mini-batches of size 6 and a fixed learning rate of $10^{-3}$.
Color jittering augmentation is applied on the RGB data.
For all the networks, we assume that
the bounding box of the person is given, thus we crop the image to center the person.
Code is made publicly available 
on the project page~\cite{projectpage}.

\subsection{Fitting a parametric body model}
\label{subsec:fitting}
While the volumetric output of BodyNet produces good quality results, for
some applications, it is important to produce a 3D surface mesh, or even a parametric 
model that can be manipulated.
Furthermore, we use the SMPL model for our evaluation.
To this end, we process the network output in two steps:
(i) we first extract the isosurface from the predicted occupancy map, (ii)
next, we optimize 
for the parameters of a deformable body model, SMPL model in our experiments, 
that fits the isosurface as well as the predicted 3D joint locations.

Formally, we define the set of 3D vertices in the isosurface mesh
that is extracted~\cite{lewiner2003mcubes} from the network output  to be $\mathbf{V}^n$. 
SMPL~\cite{smpl2015} is a statistical 
model where the location of each vertex is given by a set 
$\mathbf{V}^s(\theta,\beta)$ that is formulated as a function
of the pose ($\theta$) and 
shape ($\beta$) parameters~\cite{smpl2015}.
Given $\mathbf{V}^n$, our 
goal is to find $\{\theta^\star, \beta^\star\}$ such that the
weighted 
Chamfer distance, i.e., the distance among the closest point correspondences between $\mathbf{V}^n$ and $\mathbf{V}^s(\theta,\beta)$ is 
minimized:
\begin{align}
\nonumber { \{ \theta^\star, \beta^\star \} } =  \underset{ \{ \theta, \beta \} }{\argmin}
&   \sum_{\mathbf{p}^n \in \mathbf{V}^n}  \min_{\mathbf{p}^s \in \mathbf{V}^s(\theta, \beta)} w^n \|\mathbf{p}^n-\mathbf{p}^s\|_2^2 + \\
&  \sum_{\mathbf{p}^s \in \mathbf{V}^s(\theta, \beta)} \min_{\mathbf{p}^n \in \mathbf{V}^n} w^n\|\mathbf{p}^n - \mathbf{p}^s\|_2^2 +  \lambda \sum_{i=1}^{J} \| \mathbf{j}^n_i - \mathbf{j}^s_i(\theta, \beta) \|_2^2 .
\label{eq:objective}
\end{align}
We find it effective to weight the closest point distances by the confidence
of the corresponding point in the isosurface which depends on the voxel predictions of our network.
We denote the weight associated with the point $p^n$ as $w^n$.
We define an additional term to measure the distance between
the predicted 3D joint locations, $\{\mathbf{j}^n_i\}_{i=1}^{J}$, 
where $J$ denotes the number of joints, and the corresponding
joint locations in the SMPL model, denoted by $\{\mathbf{j}^s_i(\theta, \beta)\}_{i=1}^{J}$.
We weight the contribution of the joints' error by a constant $\lambda$ (empirically set to $5$ in our experiments)
since $J$ is very small
(e.g., 16) compared to the number of vertices (e.g., 6890).
In Sec.~\ref{sec:experiments}, we show the benefits of fitting to voxel predictions
compared to our baseline of fitting to 2D and 3D joints, and to 2D segmentation,
i.e., to the inputs of the shape network.

We optimize for Eq.~(\ref{eq:objective}) in an iterative manner
where we update the correspondences at each iteration. We use Powell's dogleg
method~\cite{dogleg} and Chumpy~\cite{chumpy} similar
to~\cite{Bogo2016smplify}.
When reconstructing the isosurface, we first apply a thresholding ($0.5$ in our experiments) to the
voxel predictions and apply the marching cubes
algorithm~\cite{lewiner2003mcubes}.  
We initialize the SMPL pose parameters to be aligned
with our 3D pose predictions and set $\beta = \vec{0}$
(where $\vec{0}$ denotes a vector of zeros).

\section{Experiments}
\label{sec:experiments}
This section presents the evaluation of BodyNet. We first describe evaluation datasets (Sec.~\ref{subsec:data})
and other methods used for comparison in this paper (Sec.~\ref{subsec:baselines}). We then evaluate
contributions of additional inputs~(Sec.~\ref{subsec:3Dshape}) and losses~(Sec.~\ref{subsec:expmultitask}).
Next, we report performance on the UP dataset (Sec.~\ref{subsec:UP}).
Finally, we demonstrate results for 3D body part segmentation~(Sec.~\ref{subsec:parts}).

\subsection{Datasets and evaluation measures}
\label{subsec:data}
\noindent\textbf{SURREAL dataset}~\cite{varol2017surreal} is a 
large-scale synthetic dataset for 3D human body shapes with ground truth labels 
for segmentation, 2D/3D pose, and SMPL body
parameters.
Given its scale and rich ground truth, we use SURREAL in this work for training
and testing. Previous work demonstrating successful use of synthetic images of people for
training visual models include \cite{barbosa,Deep3DPose,synthetic_cohenor}. 
Given the SMPL shape and pose parameters, we compute
the ground truth 3D mesh. We use the standard train split~\cite{varol2017surreal}.
For testing, we use the middle frame of the middle clip of each test
sequence, which makes a total of $507$ images. We observed that testing
on the full test set of $12,528$ images yield similar results.
To evaluate the quality of our shape predictions for difficult cases, we define two subsets with extreme body shapes, similar to
what is done for example in optical flow~\cite{Butler:ECCV:2012}. 
We compute the surface distance between
the average shape ($\beta=\vec{0}$) given the ground truth pose and the true shape.
We take the $10^{th}$ ({\em s10}) and $20^{th}$ ({\em s20}) percentile
of this distance distribution that represent the meshes with extreme body shapes.

\noindent\textbf{Unite the People dataset} (UP)~\cite{lassner2017up} is a recent collection
of multiple datasets (e.g., MPII~\cite{andriluka14mpii}, LSP~\cite{johnson2010lsp}) 
providing additional annotations for each image. 
The annotations include 2D pose with 91 keypoints, 31 body part segments,
and 3D SMPL models. The ground truth is acquired in a semi-automatic way and is therefore
imprecise.
We evaluate our 3D body shape
estimations on this dataset. 
We report errors on two different subsets of the test set where 
2D segmentations as well as pseudo 3D ground truth are available. We use notation
T1 for images from the LSP subset~\cite{lassner2017up}, and T2 for images used by~\cite{tan2017bmvc}.

\noindent\textbf{3D shape evaluation.} 
We evaluate body shape estimation with different measures.
Given the ground truth and our predicted volumetric representation, we measure the intersection over union
directly on the voxel grid, i.e., voxel IOU. We further assess the quality of the projected silhouette
to enable comparison with~\cite{lassner2017up,tan2017bmvc,hmrKanazawa17}.
We report the intersection over union (silhouette IOU), F1-score computed for foreground
pixels, and global accuracy (ratio of correctly predicted foreground and background pixels).
We evaluate the quality of the fitted SMPL model
by measuring the average error in millimeters between the corresponding vertices
in the fit and ground truth mesh (surface error).
We also report the average error between
the corresponding $91$ landmarks defined for the UP dataset~\cite{lassner2017up}.
We assume the depth of the root joint and the focal length to be known to transform
the volumetric representation into a metric space.

\subsection{Alternative methods}
\label{subsec:baselines}
We demonstrate advantages of BodyNet by comparing it to 
alternative methods.
BodyNet makes use of 2D/3D pose estimation and 2D segmentation.
We define alternative methods in terms of the same components combined differently. 

\begin{figure}[!hb]
    \centering
    \includegraphics[width=0.97\linewidth]{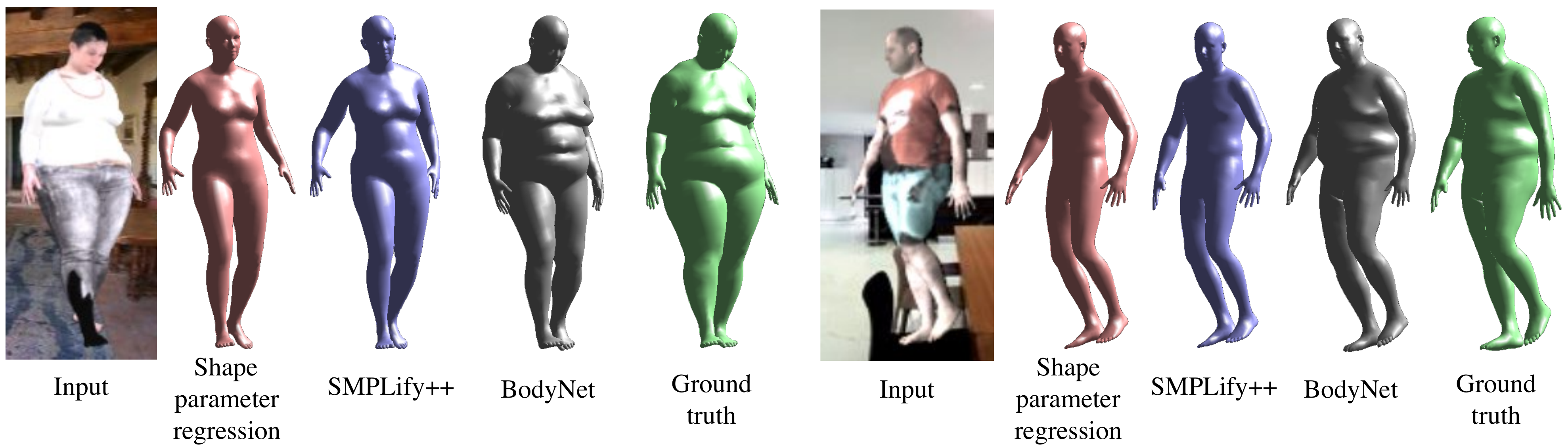}
    \caption{SMPL fit on BodyNet predictions compared with other methods.
        While shape parameter regression and the fitting only to BodyNet inputs (SMPLify++)
        produce shapes close to average,
        BodyNet learns how the true shape observed in the image deviates from the average
        deformable shape model. Examples taken from the test subset {\em s10} of SURREAL dataset with extreme shapes.}
    \label{fig:shape}
\end{figure}

\noindent{\bf SMPLify++.}
Lassner~\etal~\cite{lassner2017up} extended SMPLify~\cite{Bogo2016smplify} with an additional
term on 2D silhouette. Here, we extend it further to enable a fair comparison with BodyNet.
We use the code from~\cite{Bogo2016smplify} and implement a
fitting objective 
with additional terms on 2D silhouette
and 3D pose besides 2D pose (see Appendix~\ref{app:sec:smplifyplusplus}). 
As shown in Tab.~\ref{table:projection}, results of SMPLify++ remain inferior to 
BodyNet despite both of them using 2D/3D pose and segmentation inputs (see Fig.~\ref{fig:shape}).

\noindent{\bf Shape parameter regression.}
To validate our volumetric representation, 
 we also implement a regression method
 by replacing the 3D shape estimation network in
Fig.~\ref{fig:pipeline} by another subnetwork directly
regressing the 10-dim.~shape parameter vector $\beta$ using L2 loss.
The network architecture corresponds to the
encoder part of the
hourglass followed by 3 additional fully connected layers
(see Appendix~\ref{app:subsec:archdetails} for details).
We recover the pose parameters $\theta$ from our 3D pose prediction
(initial attempts to regress $\theta$ together with $\beta$ gave worse results).
Tab.~\ref{table:projection} demonstrates inferior performance of the $\beta$ regression network
that often produces average body shapes (see Fig.~\ref{fig:shape}).
In contrast, BodyNet results in better SMPL fitting due to the accurate volumetric representation.

\begin{table}[t]
    \centering
    \caption{Performance on the 
        SURREAL dataset using alternative combinations of intermediate representations at the input.
    }
    \resizebox{.77\linewidth}{!}{
        \begin{tabular}{lc@{\hspace{.2in}}c}
            \toprule
            & voxel IOU (\%)  & SMPL surface error (mm) \\
            \midrule
            2D pose                                           & 47.7 & 80.9 \\
            RGB                                                 & 51.8 & 79.1 \\
            Segm                                               & 54.6 &  79.1 \\
            3D pose                                           & 56.3 & 74.5\\
            Segm + 3D pose                              & 56.4 & 74.0 \\
            RGB + 2D pose + Segm + 3D pose   & \textbf{58.1} & \textbf{73.6} \\
            \bottomrule
        \end{tabular}
    }
    \label{table:inputs}
\end{table}
\begin{figure}[t]
    \centering
    \includegraphics[width=0.98\linewidth]{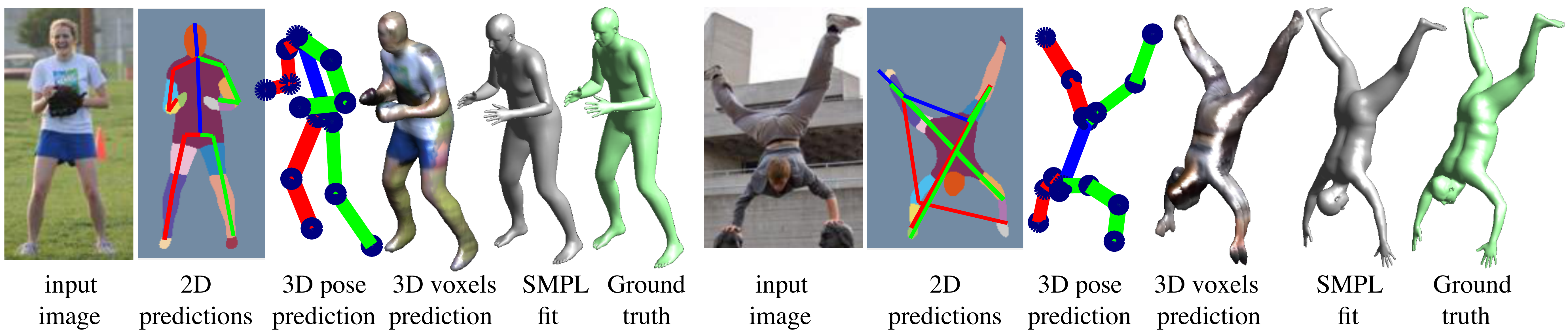}
    \caption{
        Our predicted 2D pose, segmentation, 3D pose, 3D volumetric shape, and SMPL model
        alignments. Our
        3D shape predictions are consistent with pose and segmentation, suggesting that
        the shape network relies on the intermediate representations. When one of the auxiliary
        tasks fails (2D pose on the right), 3D shape can still be recovered with the help of the other cues.
    }
    \label{fig:results}
\end{figure}
%

\subsection{Effect of additional inputs}
\label{subsec:3Dshape}
We first motivate our proposed architecture by evaluating
performance of 3D shape estimation in the SURREAL dataset
using alternative inputs (see Tab.~\ref{table:inputs}).
When only using one input, 3D pose network, which is already trained with 
additional 2D pose and segmentation inputs, performs best. 
We observe improvements as more cues, specifically 3D cues are added. 
We also note that intermediate representations in terms of 3D pose and 2D segmentation outperform RGB. 
Adding RGB to the intermediate representations further improves shape results on SURREAL. 
Fig.~\ref{fig:results} illustrates intermediate predictions as well as the final 3D shape output. 
Based on results in Tab.~\ref{table:inputs}, we choose to use all intermediate
representations as parts of our full network that we call BodyNet. 

\begin{table}
\centering
\caption{Volumetric prediction on SURREAL with different versions of our model compared
	to alternative methods. Note that lines 2-10 use same modalities (i.e., 2D/3D pose, 2D segmentation).
	The evaluation is made on the SMPL model fit to our voxel outputs. The average SMPL surface error
	decreases with the addition of the proposed components.}
\resizebox{.9\linewidth}{!}{
	\begin{tabular}{r@{\hspace{.2in}}lc@{\hspace{.2in}}c@{\hspace{.2in}}c}
		\toprule
		&& full & {\em s20} & {\em s10} \\
		\midrule
		1. &Tung \etal~\cite{tung2017selfsupervised} \quad  (using GT 2D pose and segmentation)   & 74.5 & - & - \\
		\midrule
		\midrule
		&\multicolumn{2}{l}{\em Alternative methods:} \\
		\midrule
		2. &SMPLify++ ($\theta$, $\beta$ optimized)  & 75.3    & 79.7 & 86.1 \\
		3. &Shape parameter regression ($\beta$ regressed, $\theta$ fixed) & 74.3  & 82.1 & 88.7 \\
		\midrule
		\midrule
		&\multicolumn{2}{l}{\em BodyNet:} \\
		\midrule
		4. &Voxels network                                                               & 73.6 & 81.1 & 86.3 \\
		5. &Voxels network with [FV] silhouette re-projection           & 69.9 & 76.3 & 81.3 \\
		6. &Voxels network with [FV+SV] silhouette re-projection     & 68.2 & 74.4 & 79.3 \\
		7. &End-to-end without intermediate tasks [FV]                    & 72.7 & 78.9 & 83.2 \\
		8. &End-to-end without intermediate tasks [FV+SV]              & 70.5 & 76.9 & 81.3 \\
		9. &End-to-end with intermediate tasks [FV]                         &  67.7 & 74.7 & 81.0 \\
		10. &End-to-end with intermediate tasks [FV+SV]  & \textbf{65.8} & \textbf{72.2} & \textbf{76.6} \\
		\bottomrule
	\end{tabular}
}
\label{table:projection}
\end{table}

\subsection{Effect of re-projection error and end-to-end multi-task training}
\label{subsec:expmultitask}
We evaluate contributions provided by additional supervision from Sec.~\ref{subsec:re-projection}-\ref{subsec:multi-task}.

\noindent\textbf{Effect of re-projection losses.}
Tab.~\ref{table:projection} (lines 4-10) provides results when the shape network is trained with and without 
re-projection losses (see also Fig.~\ref{fig:projection}). The
voxels network without
any additional loss already outperforms the baselines described in Sec.~\ref{subsec:baselines}.
When trained with re-projection losses,
we observe increasing performance both with single-view constraints, i.e., front view (FV), and
multi-view, i.e., front and side views (FV+SV). The multi-view re-projection loss
puts more importance on the body surface resulting in a better SMPL fit.

\noindent\textbf{Effect of intermediate losses.}
Tab.~\ref{table:projection} (lines 7-10) presents experimental evaluation of the proposed
intermediate supervision. Here, we first compare the end-to-end network fine-tuned jointly
with auxiliary tasks (lines 9-10) to the networks trained independently from the fixed representations
(lines 4-6).
Comparison of results on lines 6 and 10 suggests that multi-task training regularizes all subnetworks
and provides better performance for 3D shape. We refer to Appendix~\ref{app:subsec:multitaskeffect}
for the performance improvements on auxiliary tasks.
To assess the contribution of intermediate losses on 2D pose, segmentation, and 3D pose,
we implement an additional baseline where we again fine-tune end-to-end, but remove the
losses on the intermediate tasks (lines 7-8).
Here, we keep only the voxels and the re-projection losses.
These networks not only forget the intermediate tasks, but are also outperformed by our base
networks without end-to-end refinement (compare lines 8 and 6).
On all the test subsets (i.e., full, {\em s20}, and {\em s10}) we observe a consistent improvement of the proposed
components against baselines. Fig.~\ref{fig:shape}
presents qualitative results and illustrates
how BodyNet successfully learns the 3D shape in extreme cases.

\noindent\textbf{Comparison to the state of the art.}
Tab.~\ref{table:projection} (lines 1,10) demonstrates a significant improvement of BodyNet compared to the recent method of Tung~\etal~\cite{tung2017selfsupervised}.
Note that \cite{tung2017selfsupervised}
relies on ground truth 2D pose and 
segmentation on the test set, while our approach is fully automatic.
Other works do not report results on the recent SURREAL dataset.

\begin{table}[t]
	\centering
	\caption{Body shape performance and comparison to the state of the art on the UP dataset.
		Unlike in SURREAL, the 3D ground truth in this dataset is imprecise.
		\textsuperscript{1}This result is
		reported in~\cite{lassner2017up}. \textsuperscript{2}This result
		is reported in~\cite{tan2017bmvc}.
	}
	\resizebox{0.9\linewidth}{!}{
				\begin{tabular}{l@{\hspace{.15in}}l@{\hspace{.2in}}c@{\hspace{.2in}}c@{\hspace{.2in}}c@{\hspace{.4in}}c@{\hspace{.2in}}c}
					\toprule
					& & \multicolumn{3}{c}{\hspace{-.3in} \em 2D metrics} &\multicolumn{2}{c}{\em 3D metrics (mm)} \\
					&                                                                                                    & Acc. (\%) & IOU & F1 & Landmarks & Surface \\ 
					\midrule
					\parbox[t]{2mm}{\multirow{6}{*}{\rotatebox[origin=c]{90}{T1}}} 
					& 3D ground truth~\cite{lassner2017up}                                        & 92.17 & - & 0.88  & 0 & 0\\
					&Decision forests~\cite{lassner2017up}                                        & 86.60 & - & 0.80 & - & - \\  
					&HMR~\cite{hmrKanazawa17}                                                        & 91.30 & - & 0.86 & - & - \\
					&SMPLify, UP-P91~\cite{lassner2017up}                                        & 90.99 & - & 0.86 & - & - \\
					&SMPLify on DeepCut~\cite{Bogo2016smplify}\textsuperscript{1}  & 91.89 & - & \textbf{0.88} & - & - \\
					&BodyNet {\em (end-to-end multi-task)}                                      & 92.75 & \textbf{0.73} & 0.84 & \textbf{83.3} & \textbf{102.5}\\
					\midrule \midrule
					\parbox[t]{2mm}{\multirow{4}{*}{\rotatebox[origin=c]{90}{T2}}} 
					&3D ground truth ~\cite{lassner2017up}\textsuperscript{2}         &  95.00 & 0.82& - & 0 & 0\\
					&Indirect learning~\cite{tan2017bmvc}                                         & \textbf{95.00} & \textbf{0.83} & - & 190.0 & - \\					
					&Direct learning~\cite{tan2017bmvc}                                           & 91.00 & 0.71 & - & 105.0 & - \\  
					&BodyNet {\em (end-to-end multi-task)}                                     & 92.97 & 0.75 & \textbf{0.86} & \textbf{69.6} & \textbf{80.1}\\
					\bottomrule
				\end{tabular}
	}
	\label{table:up}
\end{table}

\subsection{Comparison to the state of the art on Unite the People}
\label{subsec:UP}
For the networks trained on the UP dataset, we initialize the weights pre-trained on
SURREAL and fine-tune with the complete
training set of UP-3D where the 2D segmentations are obtained from the provided
3D SMPL fits \cite{lassner2017up}.
We show results of BodyNet trained end-to-end with multi-view re-projection loss. 
We provide quantitative evaluation of our method in Tab.~\ref{table:up} and compare to recent
approaches~\cite{tan2017bmvc,hmrKanazawa17,lassner2017up}.
We note that some works only report 2D metrics measuring how well the 3D shape
is aligned with the manually annotated segmentation.
The ground truth is a noisy 
estimate obtained
in a semi-automatic way~\cite{lassner2017up}, whose projection is mostly accurate but not its depth. 
While our results are on par with previous approaches on 2D metrics, we note that
the provided manual segmentations and the 3D SMPL fits~\cite{lassner2017up} are noisy
and affect both the training and the evaluation
\cite{Guler2018DensePose}. Therefore, we also provide a large set of visual results in Appendices~\ref{app:sec:qualitative}, \ref{app:sec:manualsegm}
to illustrate our competitive 3D estimation quality. 
On 3D metrics, our method significantly outperforms both direct and indirect learning of \cite{tan2017bmvc}.
We also provide qualitative results in Fig.~\ref{fig:results} where we show both the intermediate outputs
and the final 3D shape predicted by our method. We observe that voxel predictions are aligned with the
3D pose predictions and  provide a robust SMPL fit. 
We refer to Appendix~\ref{app:sec:manualsegm} for an analysis on the type of segmentation
used as re-projection supervision.
%
\subsection{3D body part segmentation}
\label{subsec:parts}
\begin{figure}[t]
	\centering
	\begin{minipage}{.47\textwidth}
		\centering
		\includegraphics[height=.4\linewidth]{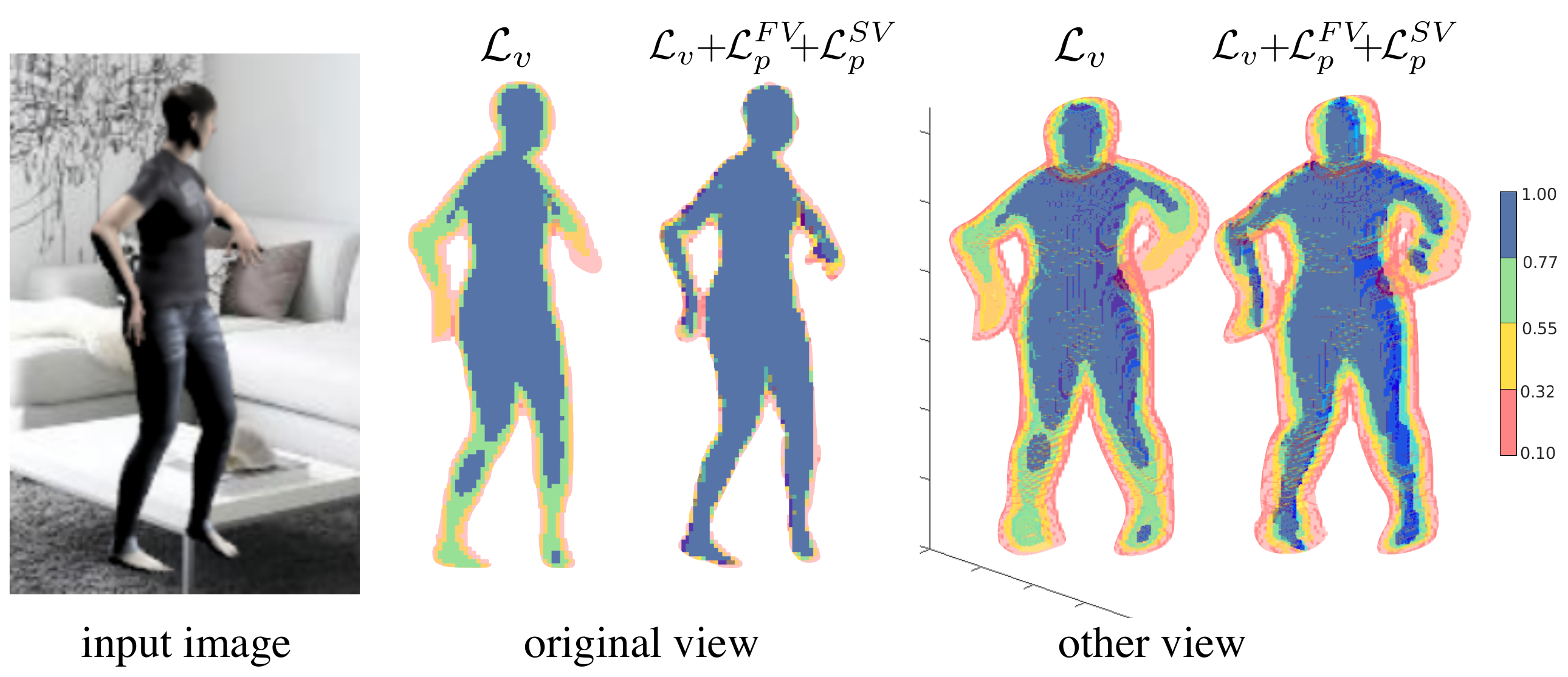}
		\captionof{figure}{Voxel predictions color-coded based on the confidence values. Notice that our
			combined 3D and re-projection loss enables our network to make more confident predictions across the 
			whole body. Example taken from SURREAL.}
		\label{fig:projection}
	\end{minipage} \quad%
	\begin{minipage}{.48\textwidth}
		\centering
		\includegraphics[height=.6\linewidth]{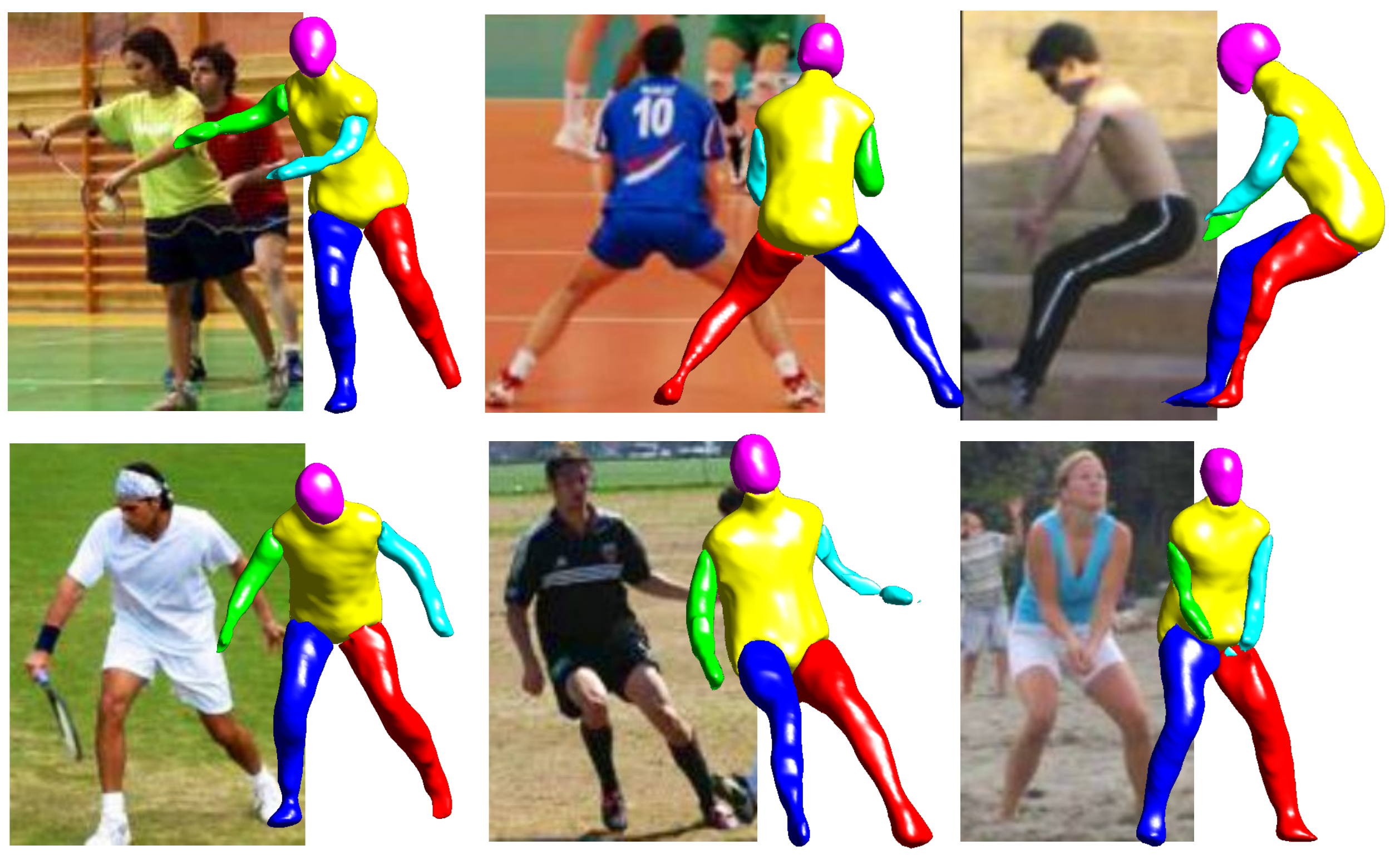}
		\captionof{figure}{BodyNet is able to directly regress volumetric body parts
			from a single image on examples from UP.}
		\label{fig:parts}
	\end{minipage}
\end{figure}

As described in Sec.~\ref{subsec:volumetric}, we extend our method to produce not only
the foreground voxels for a human body, but also the 3D part labeling. We report
quantitative results on SURREAL in Tab.~\ref{tab:parts} where accurate ground truth is available.
When the parts are combined, the foreground IOU becomes 58.9 which is comparable to 58.1
reported in Tab.~\ref{table:inputs}.
We provide
qualitative results in Fig.~\ref{fig:parts} on the UP dataset where the
parts network is only trained on SURREAL. To the best of our knowledge,
we present the first method for 3D body part labeling from a single image with an end-to-end approach.
We infer volumetric body parts directly with a network without iterative fitting of a deformable model
and obtain successful results.
Performance-wise BodyNet can produce foreground and per-limb voxels in 0.28s and 0.58s per image, respectively, using modern GPUs.
\begin{table}[t]
	\centering
	\caption{3D body part segmentation performance measured per part on SURREAL. The articulated and small limbs appear more difficult than torso.
	}
	\resizebox{.99\linewidth}{!}{
		\begin{tabular}{c@{\hspace{.15in}}c@{\hspace{.15in}}c@{\hspace{.15in}}c@{\hspace{.15in}}c@{\hspace{.15in}}c@{\hspace{.15in}}c@{\hspace{.15in}}c@{\hspace{.15in}}c}
			\toprule
			& Head & Torso & Left arm & Right arm & Left leg & Right leg & Background & Foreground \\
			\midrule
			Voxel IOU (\%) &  49.8   & 67.9    & 29.6  &  28.3 &   46.3 & 46.3 & 99.1 & 58.9\\
			\bottomrule
		\end{tabular}
	}
	\label{tab:parts}
\end{table}
\section{Conclusion}
\label{sec:conclusion}

We have presented BodyNet, a fully automatic end-to-end multi-task network architecture
that predicts the 3D human body shape from a single image. We have shown that
joint training with intermediate tasks significantly improves the results.
We have also demonstrated that the volumetric regression together with a multi-view re-projection loss
is effective for representing human bodies. 
Moreover, with this flexible representation, our framework allows us
to extend our approach
to demonstrate impressive results on 3D body part segmentation from a single image. We believe that
BodyNet can provide a trainable building block for future methods that make use of 3D body information,
such as virtual cloth-change. Furthermore, we believe exploring the limits of using only intermediate
representations is an interesting research direction for 3D tasks where acquiring training data is impractical. Another future
direction is to study the 3D body shape under clothing. Volumetric representation can potentially capture such additional geometry if 
training data is provided.

\bigskip
\noindent\textbf{Acknowledgements.}
This work was supported in part by Adobe Research, 
ERC grants \mbox{ACTIVIA} and \mbox{ALLEGRO}, the MSR-Inria joint lab, 
the Alexander von Humbolt Foundation, the Louis Vuitton ENS Chair on Artificial Intelligence, DGA project DRAAF,
an Amazon academic research award, and an Intel gift.

\bibliographystyle{splncs}
\bibliography{references}

\clearpage
\renewcommand{\thefigure}{A.\arabic{figure}} 
\setcounter{figure}{0} 
\renewcommand{\thetable}{A.\arabic{table}}
\setcounter{table}{0} 
\appendix

\title{BodyNet: Volumetric Inference of \\ 3D Human Body Shapes \\ Supplemental Material}

\titlerunning{BodyNet: Volumetric Inference of 3D Human Body Shapes}

\authorrunning{Varol, Ceylan, Russell, Yang, Yumer, Laptev, Schmid}

\author{G\"{u}l Varol\textsuperscript{$1$,*} \qquad  Duygu Ceylan\textsuperscript{$2$}
    \qquad  Bryan Russell\textsuperscript{$2$} \qquad Jimei Yang\textsuperscript{$2$} \\ 
    Ersin Yumer\textsuperscript{$2$,\ddag} \qquad  Ivan Laptev\textsuperscript{$1$,*} \qquad
      Cordelia Schmid\textsuperscript{$1$,\dag}}
\institute{\textsuperscript{$1$}Inria, France \qquad \qquad \textsuperscript{$2$}Adobe Research, USA}

\footnotetext[1]{\'{E}cole normale sup\'{e}rieure, Inria, CNRS, 
    PSL Research University, Paris, France}
\footnotetext[2]{Univ. Grenoble Alpes, Inria, CNRS, INPG, LJK, Grenoble, France}
\footnotetext[3]{Currently at Argo AI, USA. This work was performed while EY was at Adobe.}

\maketitle
\section{Qualitative analysis}
\label{app:sec:qualitative}
\subsection{Volumetric shape results}
\begin{figure}
	\centering
	\includegraphics[width=0.99\linewidth]{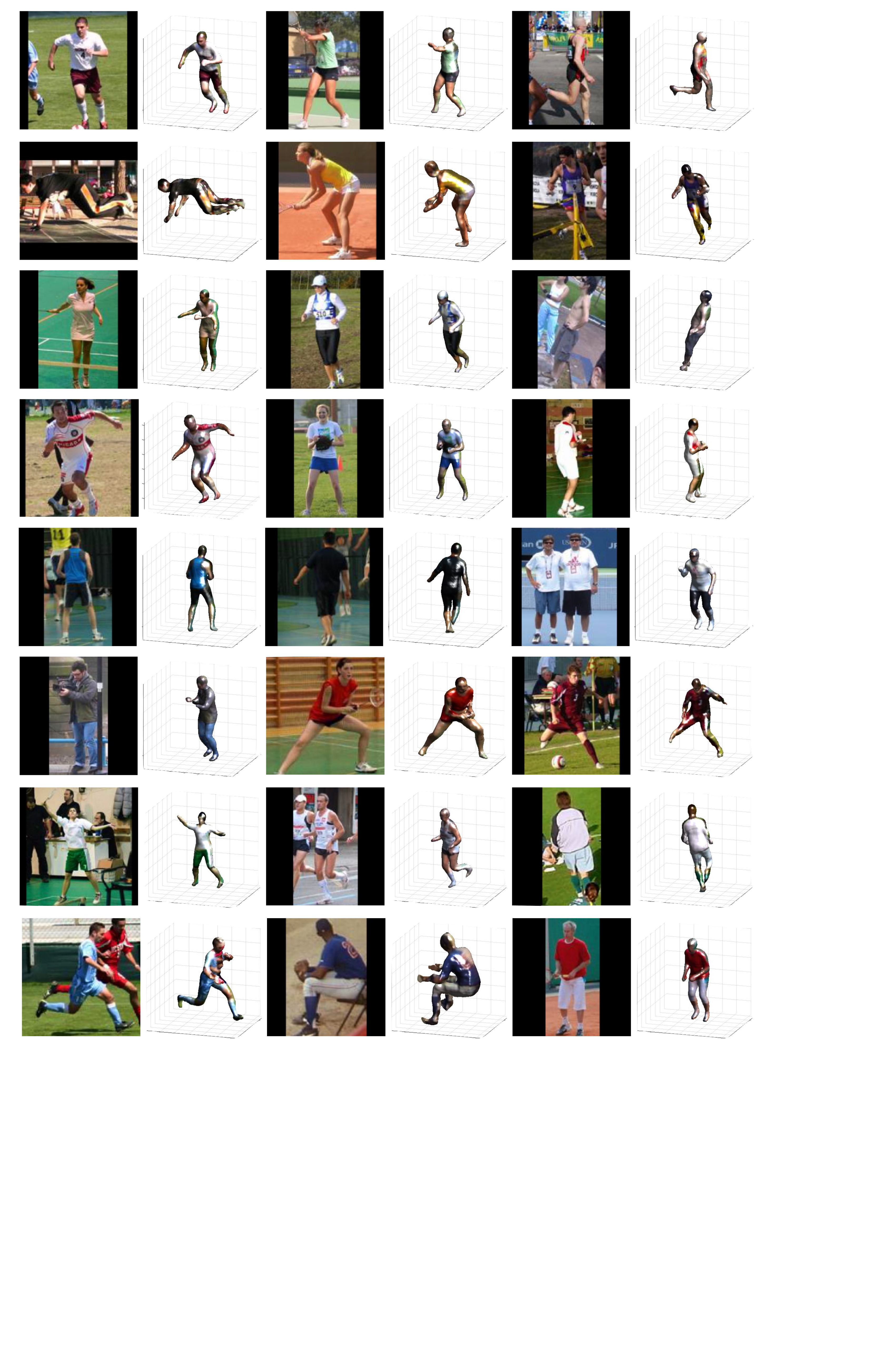}
	\caption{Qualitative results of our volumetric shape predictions on UP.}
	\label{fig:voxels}
\end{figure}

We illustrate additional examples of BodyNet output in Fig.~\ref{fig:voxels}
and in the video available in the project page~\cite{projectpage}.
We show original RGB images with corresponding predictions of 3D volumetric body shapes.
For the visualization we threshold the real-valued 3D output of the BodyNet using 0.5 as threshold
and show the fitted surface~\cite{lewiner2003mcubes}.
The texture on reconstructed bodies is automatically segmented and mapped from original images.
We also show additional examples of SMPL fits and 3D body part segmentations.
For the part segmentation, each voxel is assigned to the part with the maximum score and an isosurface is
shown per body part.
Results are shown for static images from the Unite the People dataset~\cite{lassner2017up}
and on a few real videos from YouTube. Notably, the method obtains temporally consistent results
even when applied to individual frames of the video (see video in the project page~\cite{projectpage} between 2:20-2:45).

\subsection{Predicted silhouettes versus manual segmentations on UP}
\label{app:sec:predsil}

\begin{figure}[bh!]
	\centering
	\includegraphics[width=0.99\linewidth]{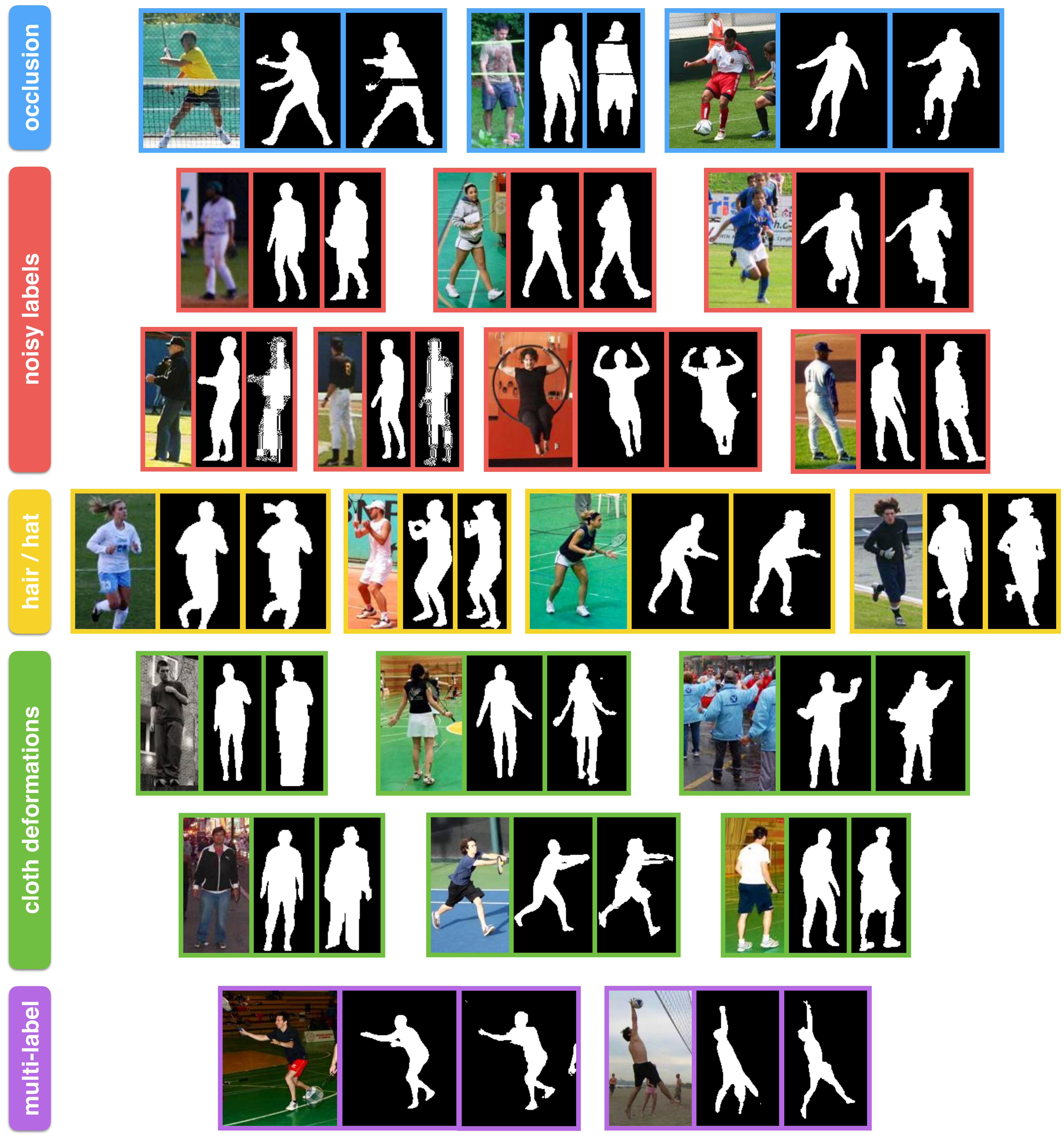}
	\caption{Projected silhouettes of our voxel predictions (middle) versus manually
		annotated segmentations (right) on the Unite the People dataset. We note that the evaluation
		is problematic due to several reasons such as occlusions and clothings, whose
		definitions are application-dependent, i.e., one might be interested in anatomic body
		or the full cloth deformation.}
	\label{fig:UP_segm}
\end{figure}

Fig.~\ref{fig:UP_segm} compares projected silhouettes of our voxel predictions (middle)
with the manually annotated segmentations (right) used as ground truth for the evaluation in Tab.~\ref{table:up}.
While our results are as expected and good, we observe frequent inconsistencies with the manual annotation
due to several reasons:
BodyNet produces a full 3D human body shape even in the case of occlusions (blue);
annotations are often imprecise (red);
the 3D prediction of cloth (green) and hair (yellow) is currently beyond this work due to the
lack of training data, we instead focus on producing the anatomic parts (e.g., two legs instead of a long
dress); finally, the labels are not always consistent in the case of multi-person images (purple).

We note that we never use manual segmentation during training as such annotations are not available
for the full UP-3D dataset. As supervision for re-projection losses we instead use the SMPL silhouettes
whose overlap with the manual segmentation is already not perfect (see Tab.~\ref{table:up}, first row).
Therefore, our performance in 2D metrics has an upper bound. Due to difficulties
with the quantitative evaluation, we mostly rely on qualitative results for the UP dataset.

\subsection{SMPL error}
\begin{figure}
    \centering
    \includegraphics[width=0.52\linewidth]{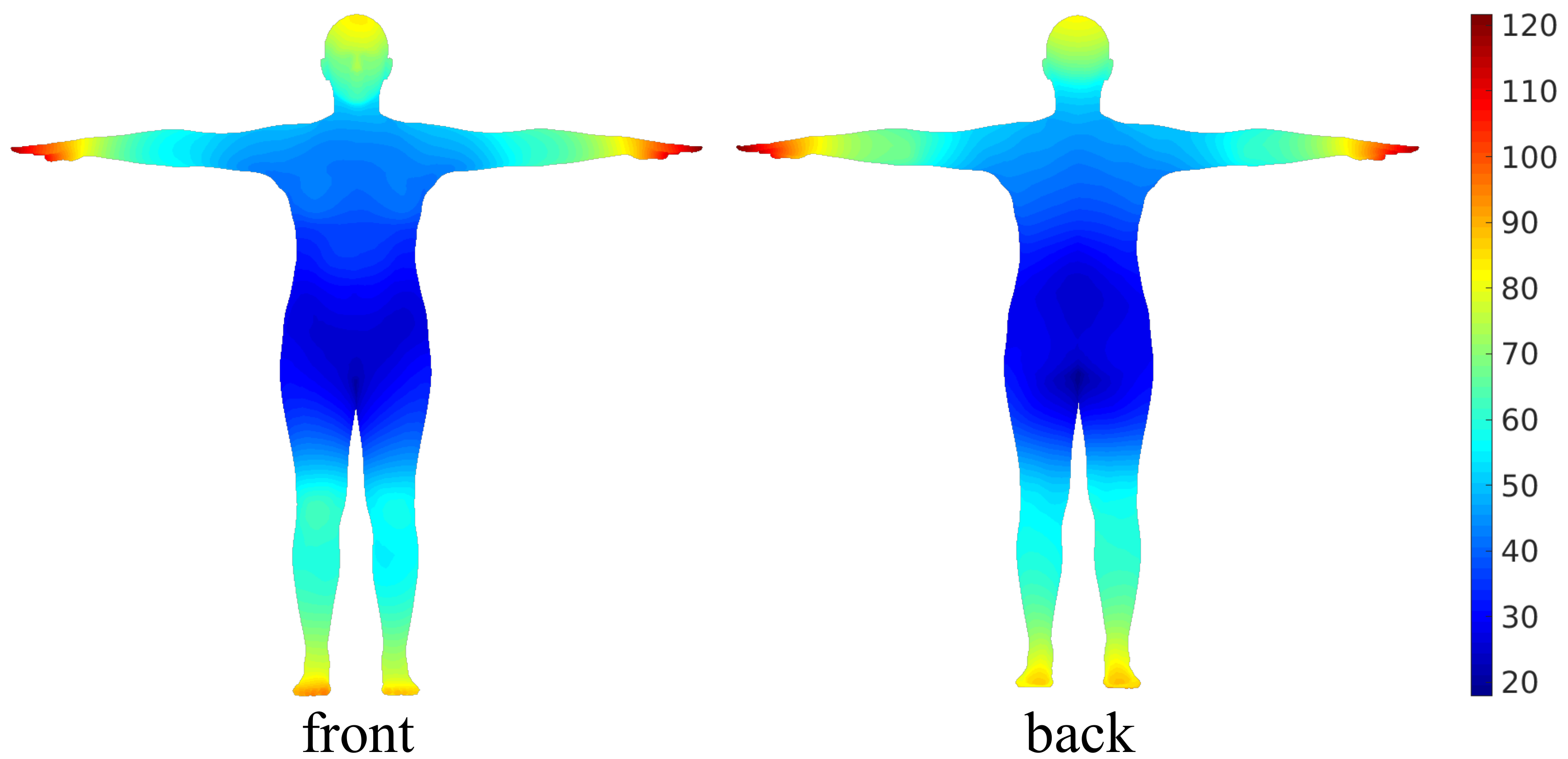}
    \caption{Per-vertex SMPL error on SURREAL. Hands and feet contribute the most to the
        surface error, followed by the other articulated body parts.}
    \label{app:fig:pervertex}
    \mbox{}\vspace{-1.5cm}\\
\end{figure}

We next investigate the quality of predictions depending on the body location.
We examine the network from Tab.~\ref{table:projection} (line 10, 65.8mm surface error)
and measure the average per-vertex error.
We visualize the color-coded SMPL surface in Fig.~\ref{app:fig:pervertex} indicating the areas with the
highest and lowest errors by the red and blue colors, respectively.
Unsurprisingly, the highest errors occur at the extremities of the body which can be explained
by the articulation and the limited resolution of the voxel grid preventing the capture of fine details.

\begin{figure}[!h]
    \begin{center}
        \includegraphics[width=0.59\linewidth]{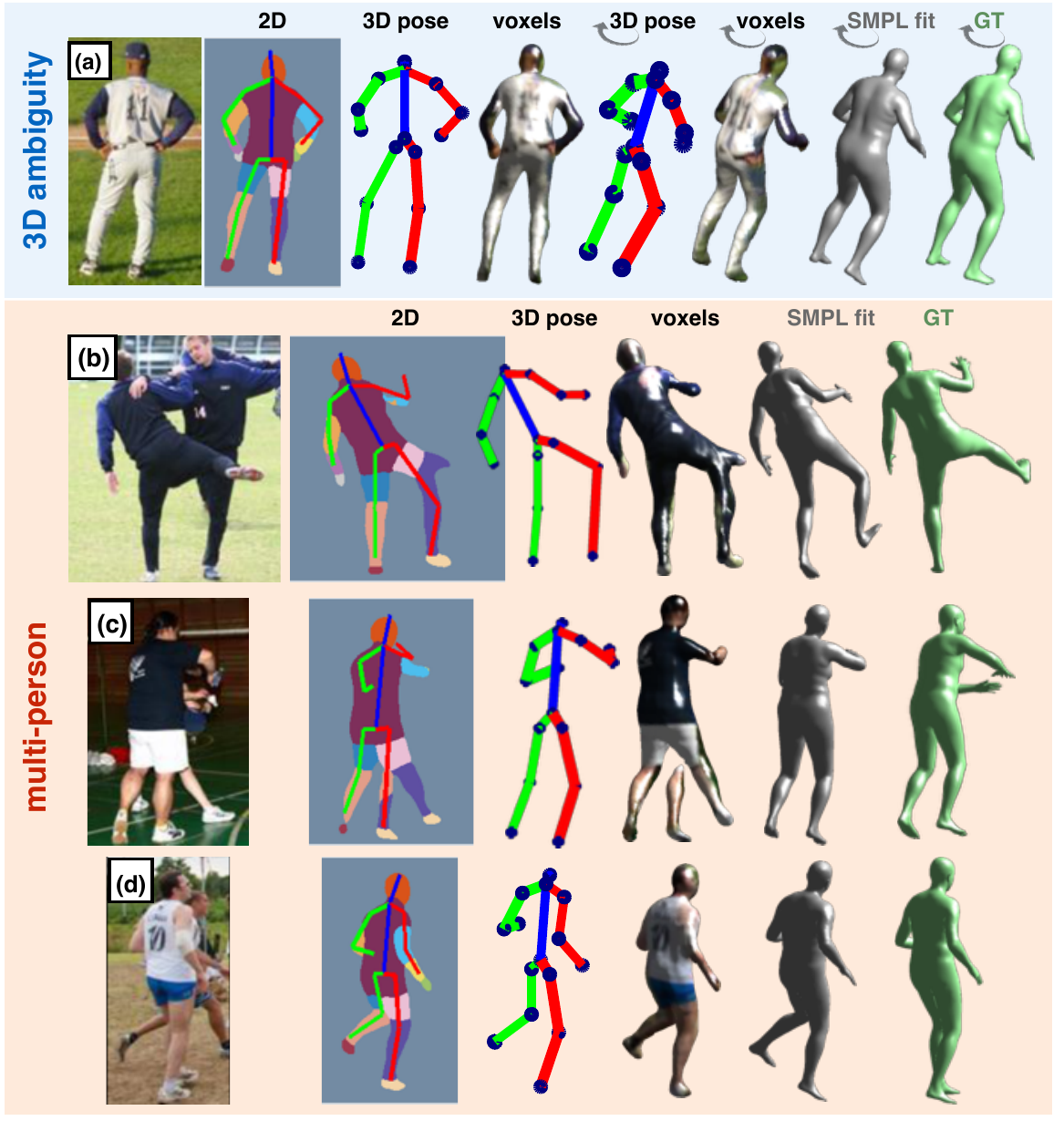}
    \end{center}
    \mbox{}\vspace{-1.3cm}\\
    \caption{Failure cases on images from UP. Arrows denote rotated views.
                Top(a): results for depth ambiguity visible with the rotated
                view. Bottom(b-d): intermediate predictions failing in a
                multi-person image. Note
                that GT is inaccurate due to the semi-automatic
                annotation protocol.}
    \label{app:fig:failures}
    \mbox{}\vspace{-1.5cm}\\
\end{figure}

\subsection{Failure modes}
\label{app:subsec:failures}
Fig.~\ref{app:fig:failures} presents failure cases on UP.
Depth ambiguity (a) and multi-person images (b-d) often cause failures of
pose estimation that propagate further to the voxel output.
Note that UP GT also has errors
and our method may learn such errors when trained on UP.

\section{Architecture details}
\label{app:subsec:archdetails}
\begin{figure}[t]
	\begin{center}
		\includegraphics[width=0.99\linewidth]{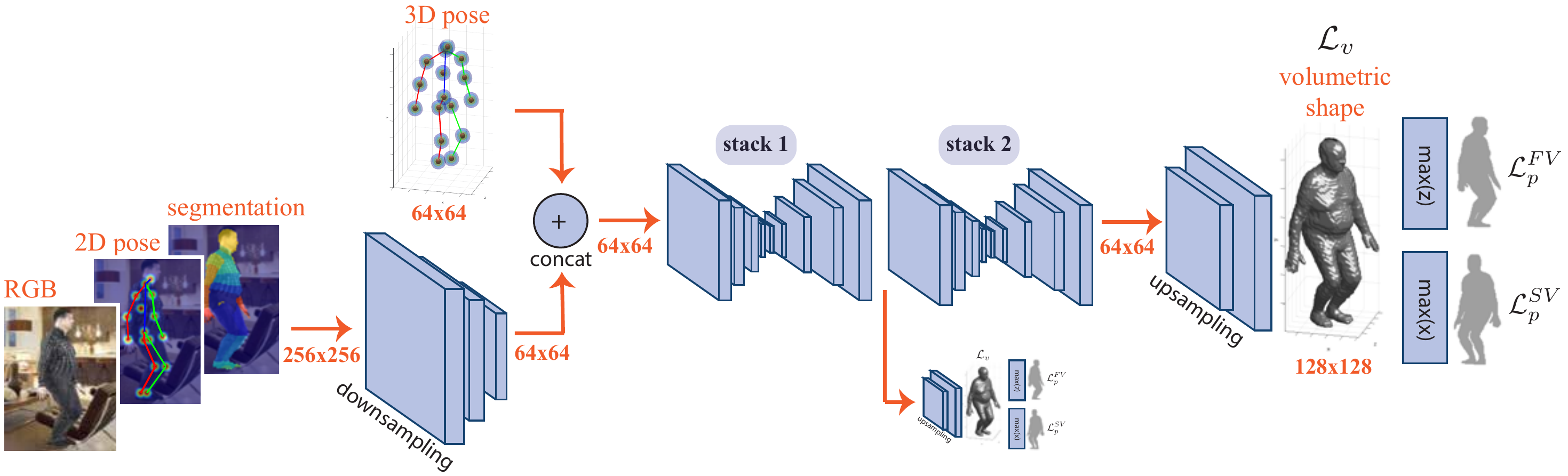}
	\end{center}
	\caption{Detailed architecture of our volumetric shape estimation subnetwork of BodyNet. The
		resolutions denoted in red refer to the \emph{spatial} resolution. See text for details.}
	\label{fig:archvolumetric}
    \mbox{}\vspace{-1.3cm}\\
\end{figure}
\begin{figure}[t]
	\begin{center}
		\includegraphics[width=0.99\linewidth]{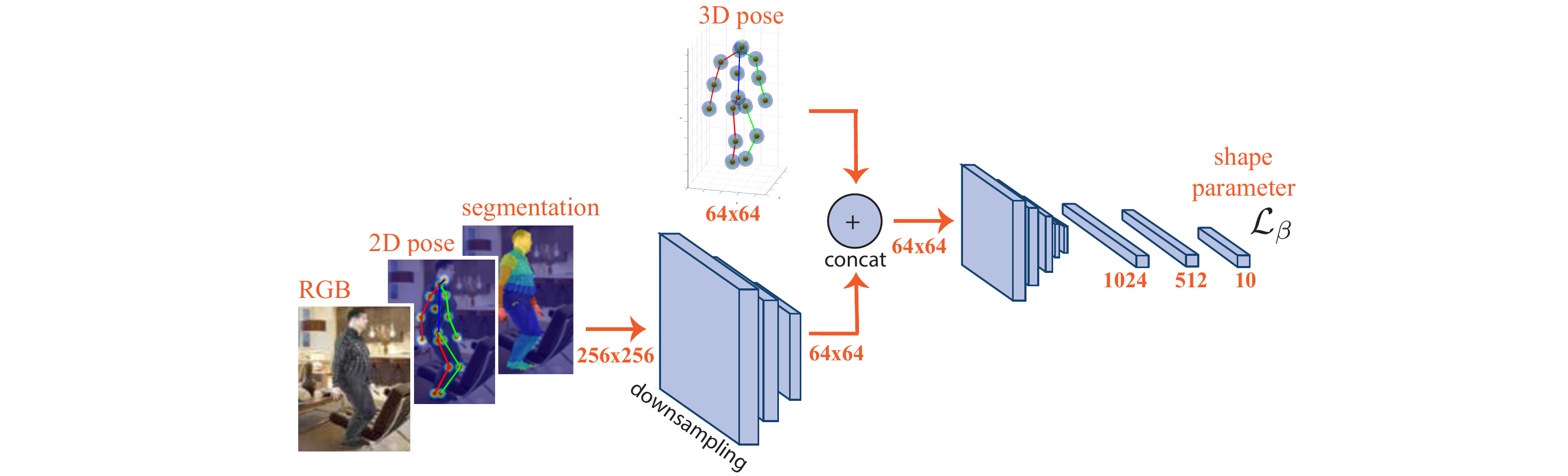}
	\end{center}
	\caption{Detailed architecture of the shape parameter regression subnetwork
		described in Sec.~\ref{subsec:baselines}.}
	\label{fig:archregression}
\end{figure}

\subsection{Volumetric shape network}
The architecture for our 3D shape estimation network is detailed in
Fig.~\ref{fig:archvolumetric}. As described in Sec.~\ref{subsec:volumetric},
this network consists of two hourglasses, each supervised by the same type of losses.
Different than the other
subnetworks in BodyNet, the input to the shape estimation network
is a combination of multiple modalities of different resolutions.
We design an architecture whose first branch operates on the concatenation of RGB ($3\times256\times256$),
2D pose  ($16\times256\times256$), and segmentation  ($15\times256\times256$)
channels as done in the original stacked hourglass network~\cite{newell2016hourglass}
where a series of convolution and pooling operations  downsample the spatial resolution with a factor of $4$.
Once the spatial resolution of this branch matches the one of the 3D pose
input, i.e., $64\times64$, we concatenate the feature maps of the first branch with 
the 3D pose heatmap channels. Note that the depth resolution of 3D pose
is treated as input channels, thus its dimensions become $304\times64\times64$
for 16 body joints and 19 depth units ($304=16\times19$). The output of the
second hourglass has again $64\times64$ spatial resolution. We use bilinear 
upsampling followed by ReLU and $3\times3$ convolutions
to obtain the output resolution of $128\times128\times128$.

\subsection{Shape parameter regression network}
We described shape parameter regression as an alternative method in Sec.~\ref{subsec:baselines}.
Fig.~\ref{fig:archregression} gives architectural details for this subnetwork. The input part of the network
is the same as in Fig.~\ref{fig:archvolumetric}. 
The output resolution at the bottleneck layer of the hourglass is $128\times4\times4$ (i.e., 2048-dim).
We vectorize this output and add 3 fully connected layers of size $fc1$(2048, 1024), $fc2$(1024, 512)
and $fc3$(512, 10) to produce the 10-dim $\beta$ vector with shape parameters of the SMPL~\cite{smpl2015}
body model. This subnetwork is trained with the L2 loss.

\subsection{3D body part segmentation network}
When extending our shape network to produce 3D body parts as described in Sec.~\ref{subsec:volumetric},
we first copy the weights of the
shape network trained without any re-projection loss (line 4 of the Tab.~\ref{table:projection}).
We first train this network for 3D body parts and then fine-tune it with the additional multi-view re-projection losses.
We apply one re-projection loss per part and per view, i.e., 7*2=14 binary cross-entropy losses
for 6 parts and 1 background, for frontal and side views. For 6 parts, we apply the $max$ operation
as in Sec.~\ref{subsec:re-projection}. For the background class, we apply the $min$ operation to approximate
orthographic projection.

\section{Performance of intermediate tasks}

\subsection{Effect of multi-task training}
\label{app:subsec:multitaskeffect}
Tab.~\ref{table:multitask} reports the results before and after
end-to-end training for 2D pose, segmentation, and 3D pose (lines 6 and 10 of Tab.~\ref{table:projection}). We report mean IOU of the 14 foreground parts (excluding the background)
as in \cite{varol2017surreal} for segmentation performance. 2D pose performance
is measured with PCKh@0.5 as in \cite{newell2016hourglass}. 
We measure the 3D pose error averaged over 16 joints in millimeters.
We report the error of our predictions against ground truth with both
of them centered at the root joint. We further assume
the depth of the root joint to be given 
in order to convert $xy$ components
of our volumetric 3D pose representation in pixel space
into metric space.
The joint training for all tasks improves both the accuracy of 3D shape estimation as well as the 
performance of all intermediate tasks.
\begin{table}[!hb]
	\centering
	\caption{Performances of intermediate tasks before and after end-to-end multi-task fine-tuning on the SURREAL dataset. All 2D pose, segmentation and 3D pose results improve with the joint training.}
	\resizebox{0.99\linewidth}{!}{
		\begin{tabular}{lc@{\hspace{.4in}}c@{\hspace{.4in}}c}
			\toprule
			                 & Segmentation   & 2D pose & 3D pose \\
			                 & mean parts IOU (\%)   & PCKh@0.5 & mean joint distance (mm) \\
			\midrule
			Independent single-task training  & 59.2        & 82.7   &  46.1 \\
			Joint multi-task training         & \textbf{69.2}        & \textbf{90.8}   &  \textbf{40.8} \\
			\bottomrule
		\end{tabular}
	}
	\label{table:multitask}
\end{table}

\begin{figure}[t]
    \begin{center}
        \includegraphics[trim={7cm, 9cm, 7cm, 9cm}, clip, width=0.19\linewidth]{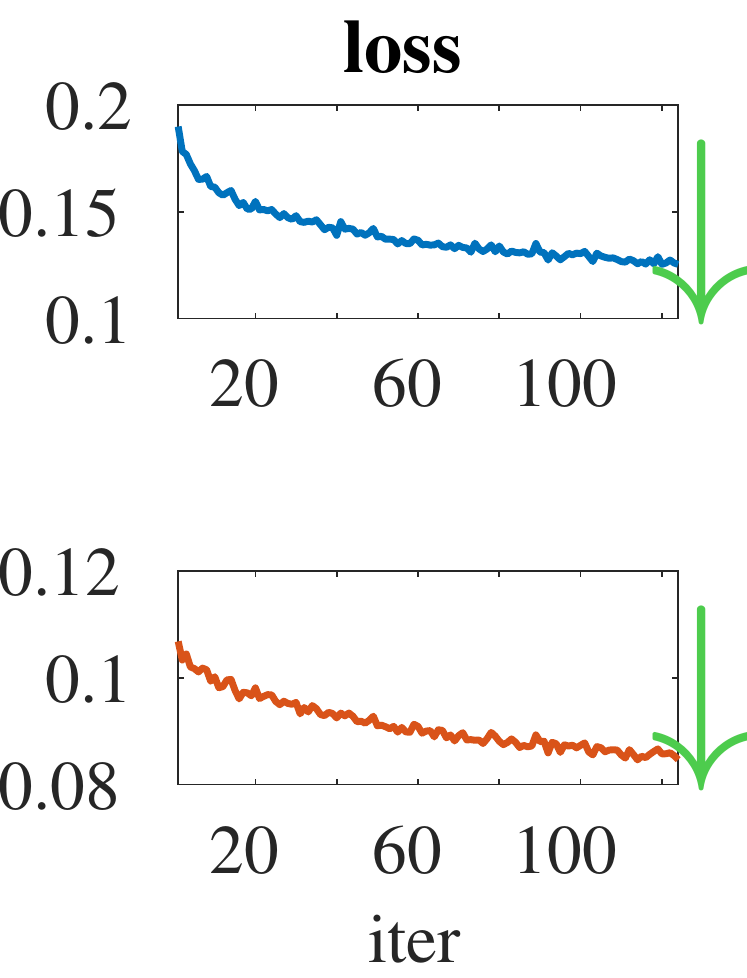}
        \includegraphics[trim={7cm, 9cm, 7cm, 9cm}, clip, width=0.19\linewidth]{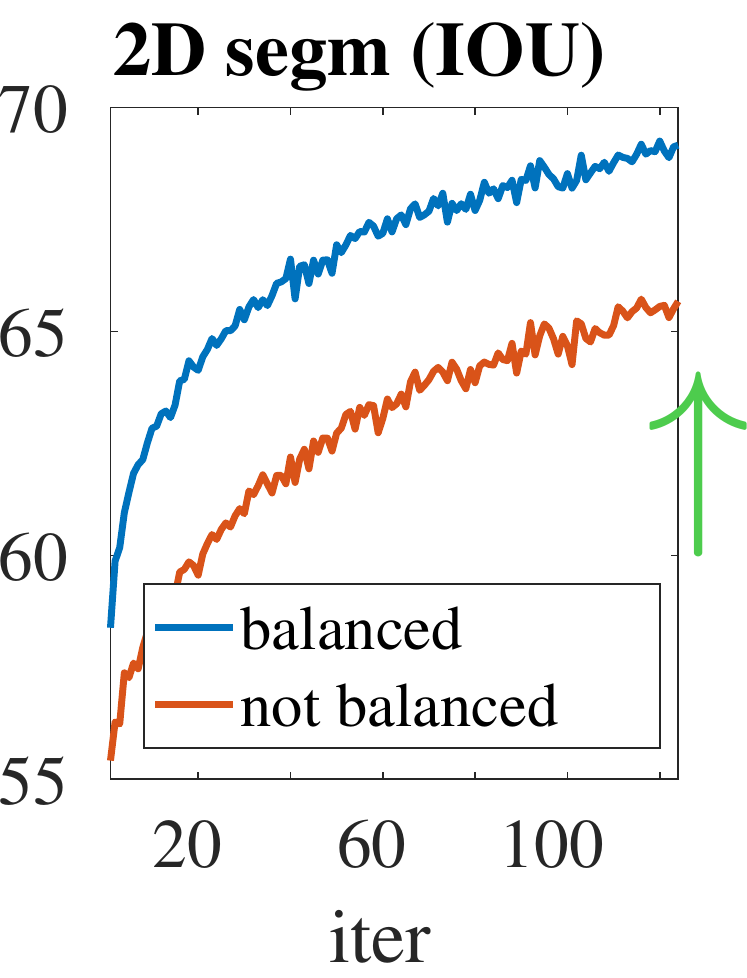}
        \includegraphics[trim={7cm, 9cm, 7cm, 9cm}, clip, width=0.19\linewidth]{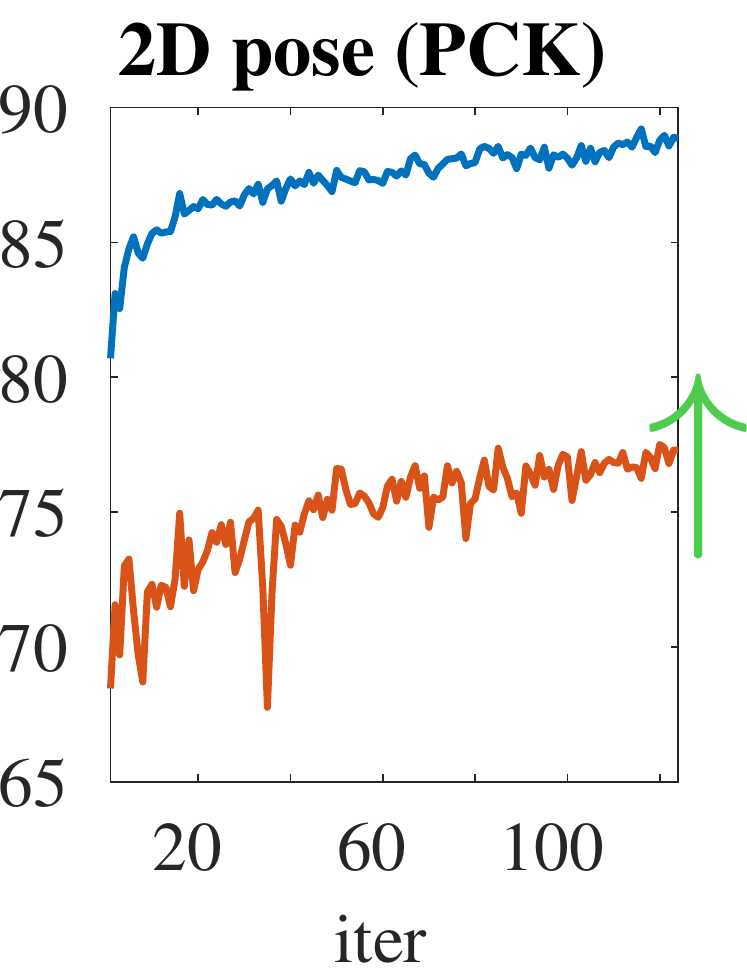}
        \includegraphics[trim={7cm, 9cm, 7cm, 9cm}, clip, width=0.19\linewidth]{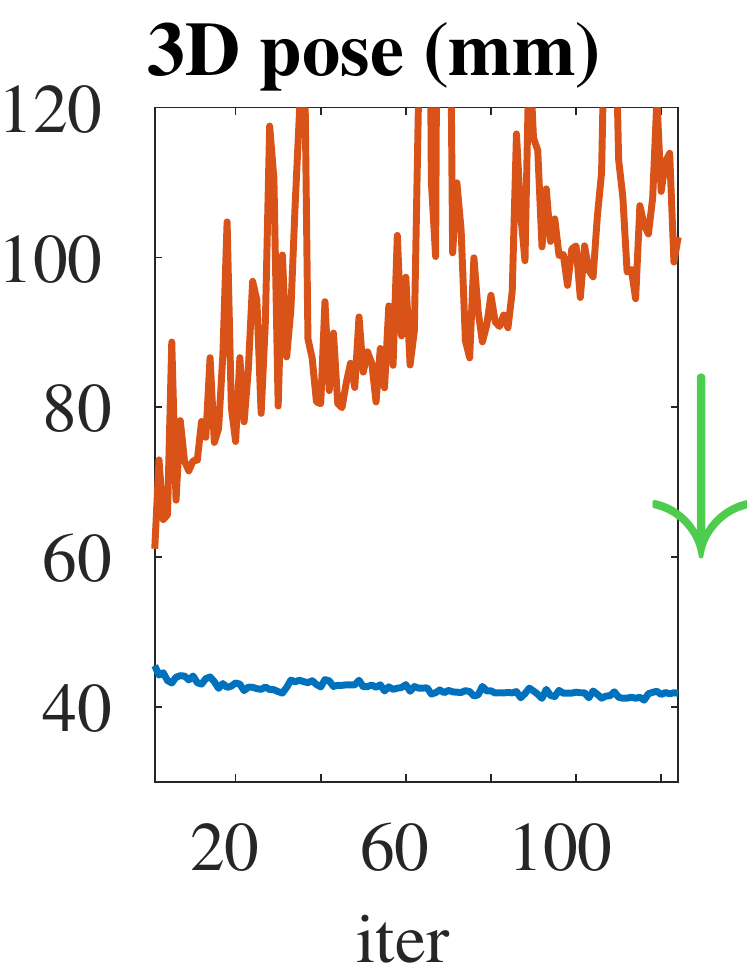}
        \includegraphics[trim={7cm, 9cm, 7cm, 9cm}, clip, width=0.19\linewidth]{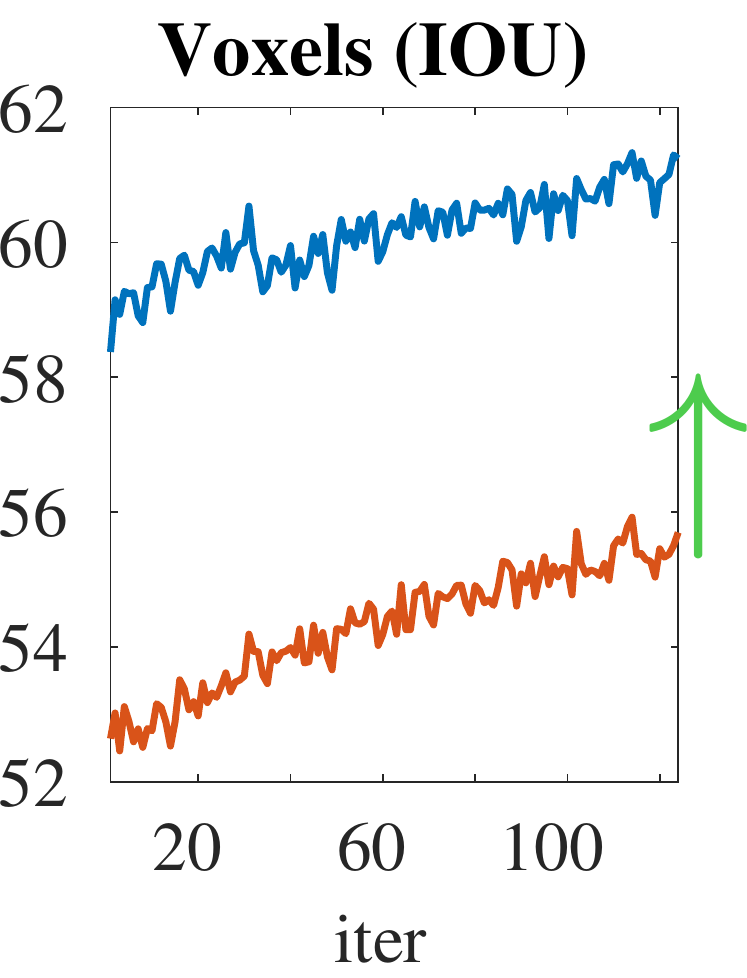}
    \end{center}
    \mbox{}\vspace{-1.3cm}\\
    \caption{Training curves with (blue) and without (red) multi-task loss balancing. Loss
        and the performance of each task are plotted at every 2K iterations. Balancing is crucial
        to equally learn all tasks, see especially 3D pose that is
        balanced with a factor $10^6$.}
    \label{app:fig:multitask}
\end{figure}
\subsection{Balancing multi-task losses}
\label{app:subsec:balancingeffect}
We set the weights in the multi-task loss by bringing the gradients of individual losses
to the same scale (see Sec.~\ref{subsec:multi-task}). For this, we set all weights to be equal
(sum to 1) and run the training for 100 iterations. We then average the gradient magnitudes
and find relative ratios to scale individual losses. In Fig.~\ref{app:fig:multitask}, we show the training curves
with and without such balancing.

\subsection{2D segmentation subnetwork on the UP dataset}
We give details on how the segmentation network that is pre-trained on SURREAL is fine-tuned on
the UP dataset. Furthermore, we report the performance and compare it to \cite{lassner2017up}.

\begin{figure}
    \begin{center}
        \includegraphics[width=0.99\linewidth]{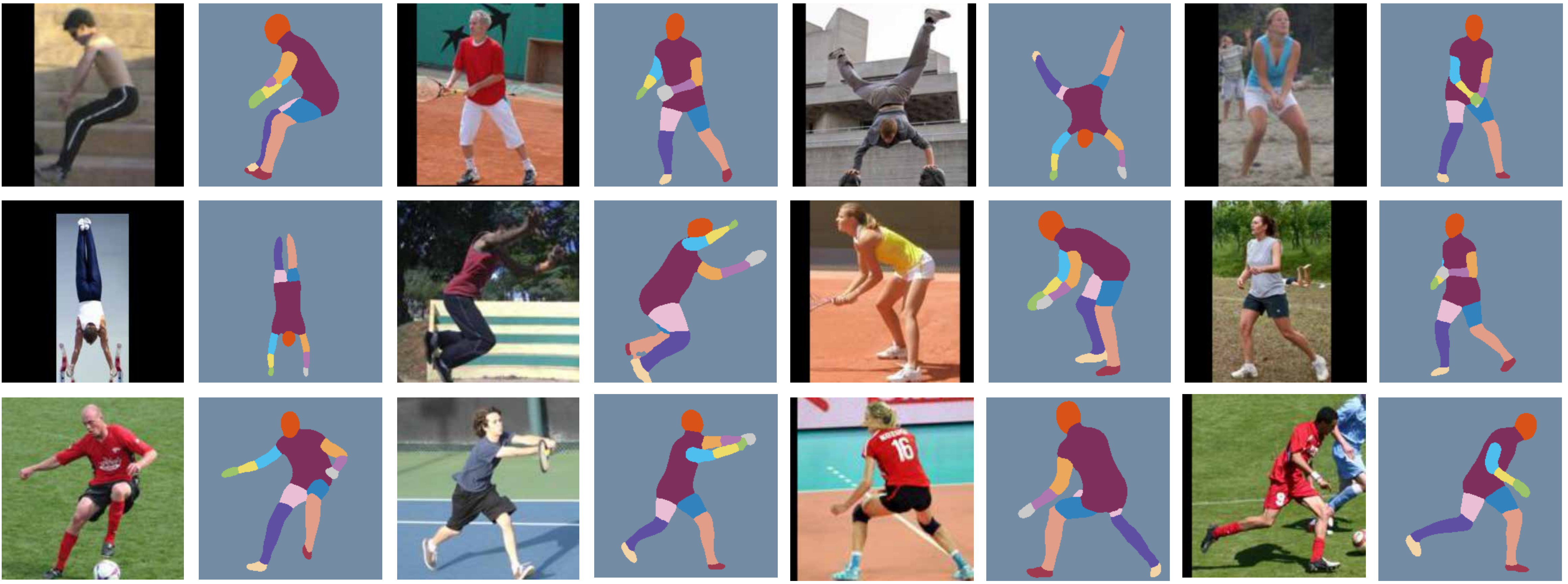}
    \end{center}
    \mbox{}\vspace{-1.3cm}\\
    \caption{Qualitative 2D body part segmentation results on the UP dataset.}
    \label{fig:UPparts}
\end{figure}

The segmentation network of BodyNet requires 15 classes (14 body parts and the background).
On the UP dataset, there are several types of segmentation annotations. The training set of
UP-3D has 5703 images with 31 part labels obtained from the projections of the automatically generated SMPL
ground truth. Manual segmentation of six body parts only exists for the LSP subset of 639 images
out of the full 1000 images with manual segmentations (not all have SMPL ground truth).
We group the 31 SMPL parts into 14, which changes the definition of some part boundaries slightly,
but are quickly learned during fine-tuning. With this strategy, we obtain 5703 training images.
Fig.~\ref{fig:UPparts} shows qualitative results for the segmentation capability of our network.
For quantitative evaluation, we use the full 1000 LSP images and group our 14 parts into 6.
We report macro F1 score that is averaged over 6 parts and the background as in \cite{lassner2017up}.
Tab.~\ref{table:UPparts} compares to other results reported in~\cite{lassner2017up}. Our subnetwork
demonstrates state-of-the-art results.

\begin{table}
    \centering
    \caption{Performance of our segmentation subnetwork on the UP dataset. See text for details.}
    \resizebox{0.8\linewidth}{!}{
        \begin{tabular}{lc}
            \toprule
            & avg macro F1 \\
            \midrule
            Trained with LSP SMPL projections \cite{lassner2017up} & 0.5628 \\
            Trained with the manual annotations \cite{lassner2017up}& 0.6046     \\
            Trained with full training (31 parts) \cite{lassner2017up} & 0.6101  \\
            Trained with full training (14 parts), pre-trained on SURREAL (ours) &  \textbf{0.6397} \\
            \bottomrule
        \end{tabular}
    }
    \label{table:UPparts}
\end{table}

\subsection{Effect of additional inputs for 3D pose}
\label{app:subsec:3Dpose}
In this section, we motivate the initial layers of the BodyNet architecture.
Specifically,  
we investigate the effect of using different input combinations of
RGB, 2D pose, and 2D segmentation for the 3D pose estimation task.
For this experiment, we do not perform end-to-end fine-tuning (similar
to Tab.~\ref{table:inputs}).
Tab.~\ref{table:input} shows the effect of gradually adding more cues at the input level and demonstrates consistent
improvements on two different datasets.
Here, we report results on both SURREAL and the widely used 3D pose benchmark
of Human3.6M dataset~\cite{h36m_pami}. 
We fine-tune our networks which are pre-trained on SURREAL
by using sequences from subjects S1, S5, S6, S7, S8, S9
and evaluate on every 64th frame of camera 2 recording subject S11 (i.e.,
{\em protocol 1} described in~\cite{rogez2017lcrnet}).

\begin{table}[b]
    \centering
    \caption{3D pose error (mm) of our 3D pose prediction network when different intermediate representations are used as input. Notice that combining all input cues yields best results, which achieves state of the art.
    }
    \resizebox{0.6\linewidth}{!}{
        \begin{tabular}{lc@{\hspace{.4in}}c}
            \toprule
            Input                 & SURREAL   & Human3.6M \\
            \midrule
            RGB                   & 49.1        & 51.6      \\
            2D pose               & 55.9         & 57.0      \\
            Segm                  & 48.1         & 58.9      \\
            2D pose + Segm        & 47.7         & 56.3     \\
            RGB + 2D pose + Segm  &\textbf{46.1} & \textbf{49.0} \\
            \midrule
            Kostrikov \& Gall ~\cite{kostrikov2014} && 115.7 \\
            Iqbal \etal~\cite{Yasin2016}            && 108.3 \\
            Rogez \& Schmid ~\cite{rogez2016nips}   && 88.1  \\
            Rogez \etal~\cite{rogez2017lcrnet}      && 53.4  \\
            \bottomrule
        \end{tabular}
    }
    \label{table:input}
\end{table}

We compare our 3D pose estimation with the state-of-the-art methods
in Tab.~\ref{table:input}.
Note that unlike others, we do not apply any rotation transformation on our output
before evaluation and we assume the depth of the root joint to be known.
While these are therefore not directly comparable,
our approach
achieves state-of-the-art performance on the Human3.6M dataset.

\section{SMPLify++ objective}
\label{app:sec:smplifyplusplus}
We described SMPLify++ as an alternative method in Sec.~\ref{subsec:baselines}. Here, we
describe the objective function to fit SMPL model
to our 2D/3D pose and 2D segmentation predictions. Given the 2D silhouette
contour predicted by the network $\mathbf{S}^n$,
our goal is to find
$\{\theta^\star, \beta^\star\}$ such that the
weighted distance among the closest point correspondences between $\mathbf{S}^n$ and $\mathbf{S}^s(\theta,\beta)$ is 
minimized:
\begin{align}
\nonumber { \{ \theta^\star, \beta^\star \} } =  \underset{ \{ \theta, \beta \} }{\argmin}
&  \sum_{\mathbf{p}^s \in \mathbf{S}^s(\theta, \beta)} \min_{\mathbf{p}^n \in \mathbf{S}^n} w^n\|\mathbf{p}^n - \mathbf{p}^s\|_2^2 + \\
&  \lambda_j \sum_{i=1}^{J} \| \mathbf{j}^n_i - \mathbf{j}^s_i(\theta, \beta) \|_2^2  + \sum_{i=1}^{J} \| \mathbf{k}^n_i - \mathbf{k}^s_i(\theta, \beta) \|_2^2 ,
\label{eq:objective2}
\end{align}
where $\mathbf{S}^s(\theta,\beta)$ is the projected silhouette of the SMPL model.
Prior to the optimization, we initialize the camera parameters with the original function from
SMPLify~\cite{Bogo2016smplify} that only uses the hip and shoulder joints for an estimate.
We use this function for initialization and further optimize the camera parameters using our
2D/3D joint correspondences. We use these camera parameters to compute the projection.
The weights $w^n$ associated to the contour point $p^n$ denote the pixel distance between
$p^n$ and its closest point (divided by the pixel threshold 10, defined by \cite{Bogo2016smplify}).

Similar to Eq. (\ref{eq:objective}),
the second
term measures the distance between
the predicted 3D joint locations, $\{\mathbf{j}^n_i\}_{i=1}^{J}$, 
where $J$ denotes the number of joints, and the corresponding
joint locations in the SMPL model, denoted by $\{\mathbf{j}^s_i(\theta, \beta)\}_{i=1}^{J}$.
Additionally, we define predicted 2D joint locations, $\{\mathbf{k}^n_i\}_{i=1}^{J}$, and
2D SMPL joint locations, $\{\mathbf{k}^s_i(\theta, \beta)\}_{i=1}^{J}$.
We set the weight $\lambda_j=100$ by visual inspection. We observe that it becomes
difficult to tune the weights with multiple objectives.
We optimize for Eq.~(\ref{eq:objective2}) in an iterative manner where we update the correspondences
at each iteration.

\section{Effect of using manual segmentations for re-projection}
\label{app:sec:manualsegm}
\begin{figure}
	\centering
	\includegraphics[width=0.99\linewidth]{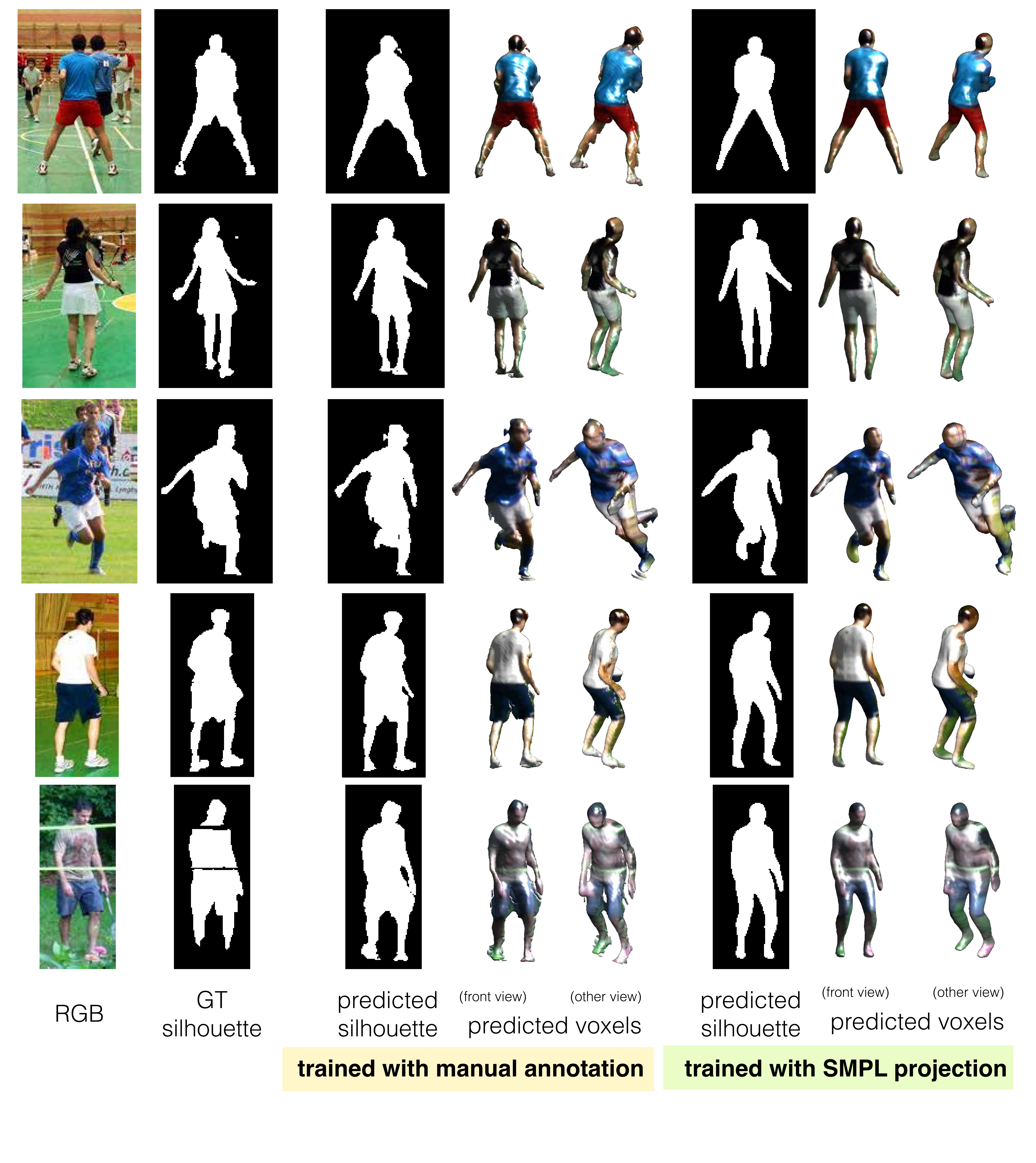}
	\caption{Using manual segmentations (M-network) versus SMPL projections (S-network)
		as re-projection supervision on the UP dataset.}
	\label{fig:UPmanual}
\end{figure}

As stated in Appendix~\ref{app:sec:predsil}, experiments in our main paper do not use the manual segmentation of the UP
dataset for training, although the evaluation on 2D metrics is performed against this ground truth.
Here we experiment with the option of using manual annotations for the front view re-projection loss (M-network)
versus the SMPL projections (S-network) as supervision.
Tab.~\ref{app:table:up2} summarizes results. We obtain significantly better aligned silhouettes
with M-network by using the manual annotations during training.
However; in this case, the volumetric supervision is not in agreement with the 2D re-projection loss.
We observe that this problem creates artifacts in the output 3D shape. Fig.~\ref{fig:UPmanual}
illustrates this effect. We show results from both M-network and S-network. Note that while the cloth boundaries
are better captured with the M-network from the front view, the output appears noisy from a rotated view.

\begin{table}
	\centering
	\caption{2D metrics on the UP dataset to compare manual segmentations (M-network) versus SMPL projections
		(S-network) as re-projection supervision.
	}
	\resizebox{0.7\linewidth}{!}{
		\begin{tabular}{l@{\hspace{.15in}}l@{\hspace{.2in}}c@{\hspace{.2in}}c@{\hspace{.2in}}c}
			\toprule
			&                                                                                                    & Acc. (\%) & IOU & F1  \\ 
			\midrule
			\parbox[t]{2mm}{\multirow{3}{*}{\rotatebox[origin=c]{90}{T1}}} 
			&SMPLify on DeepCut~\cite{Bogo2016smplify}\textsuperscript{1}   & 91.89 & - & 0.88  \\
			&S-network {\em (SMPL projections)}                                              & 92.75 & 0.73 & 0.84 \\
			&M-network {\em (manual segmentations)}                                     & \textbf{94.67} & \textbf{0.80} & \textbf{0.89}\\
			\midrule \midrule
			\parbox[t]{2mm}{\multirow{3}{*}{\rotatebox[origin=c]{90}{T2}}} 
			&Indirect learning~\cite{tan2017bmvc}                                              & 95.00 & \textbf{0.83} & - \\				
			&S-network {\em (SMPL projections)}                                              & 92.97 & 0.75 & 0.86\\
			&M-network {\em (manual segmentations)}                                     & \textbf{95.11} & 0.82 & \textbf{0.90}\\
			\bottomrule
		\end{tabular}
	}
	\label{app:table:up2}
\end{table}

\end{document}